\documentclass[10pt,twocolumn,letterpaper]{article}

\usepackage[pagenumbers]{cvpr} 

\usepackage{bm}

\renewcommand{\paragraph}[1]{\vspace{.5em}\noindent\textbf{#1.}}

\usepackage{xspace}
\usepackage{listings}
\DeclareTextSymbol{\textquotedbl}{OT1}{34}
\usepackage{tabularx}
\usepackage{threeparttable}
\AtBeginDocument{%
  \crefname{lstlisting}{Listing}{Listings}%
  \Crefname{lstlisting}{Listing}{Listings}%
  \crefname{table}{Table}{Tables}%
  \Crefname{table}{Table}{Tables}%
}
\definecolor{lstaccent}{RGB}{31,73,125}
\lstdefinestyle{prompt}{moredelim=[l][\color{lstaccent}\bfseries]{=====}}
\lstdefinestyle{kb}{emph={type,object_class,condition,strategies,knowledge,open_quantities,cross_references},emphstyle=\color{lstaccent}\bfseries,emph={[2]demonstrated,inferred,untested},emphstyle={[2]\color{gray}\itshape}}

\usepackage{algorithm}
\usepackage{algpseudocode}

\newcommand{\kb}{knowledge base\xspace}
\usepackage{placeins}

\definecolor{cvprblue}{rgb}{0.21,0.49,0.74}
\usepackage[pagebackref,breaklinks,colorlinks,allcolors=cvprblue]{hyperref}

\def\paperID{*****} 
\def\confName{CVPR}
\def\confYear{2027}

\title{RoboAuto: Autonomous Robot Program Search through \\Multi-Level Physical Understanding}

\author{Buzhao Liu$^{*}$ \quad Jianing Ye$^{*}$ \quad Xiu Yuan$^{*}$ \quad Yangtong Feng \quad Chengsong Huang\\
Tyler Dinh \quad Yevgeniy Vorobeychik \quad Ning Zhang \quad Chongjie Zhang\\
Washington University in St. Louis
}

\begin{document}
\twocolumn[{%
\renewcommand\twocolumn[1][]{#1}%
\maketitle
\begin{center}
  \centering
  \captionsetup{type=figure}
  \includegraphics[width=\textwidth]{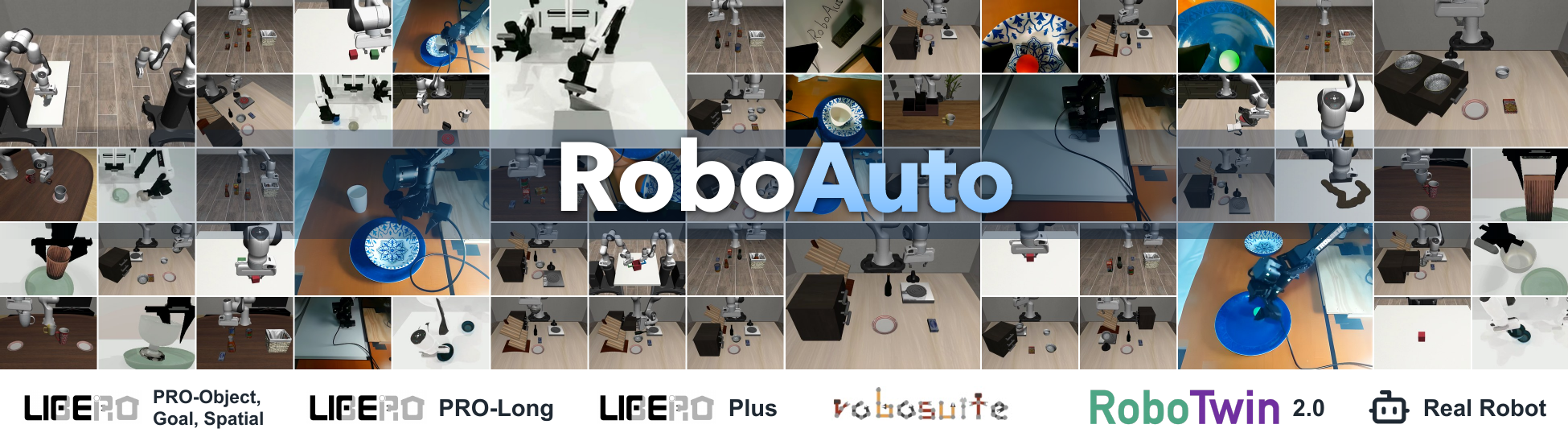}
  \captionof{figure}{\textbf{RoboAuto searches for robot programs through physical understanding.} Frames from RoboAuto's programs on LIBERO-PRO, LIBERO-PRO Long, LIBERO-Plus, Robosuite, RoboTwin~2.0, and a real robot.}
  \label{fig:first}
\end{center}%
\vspace{0.5em}
}]

{\renewcommand{\thefootnote}{\fnsymbol{footnote}}\footnotetext[1]{Equal contribution.}}

\crefname{appendix}{Appendix}{Appendices}
\Crefname{appendix}{Appendix}{Appendices}

\begin{abstract}

Recently, LLM-based autonomous agents have demonstrated strong capabilities in iteratively generating, evaluating, and refining executable solutions.
In robotic manipulation, Code-as-Policy enables these agents to search over modular and editable robot programs.
However, effective program search requires physical understanding: agents must understand how objects can be manipulated, how scene geometry constrains actions, and why interactions succeed or fail.
Existing methods rely mainly on execution feedback for program refinement, often leaving these physical questions unresolved.
To address this limitation, we introduce RoboAuto, a framework for robot program search that explicitly organizes the physical understanding at the object, scene, and interaction levels through three complementary components.
First, the hierarchical knowledge base combines executable skills with object-centric knowledge of affordances, spatial relations, and action effects, which can be learned from embodied data and online task solving.
Second, scene grounding connects this knowledge to the current environment by identifying relevant objects and resolving action-critical quantities.
Third, multi-stage planning with self-designed experiments checks whether each stage produces its expected physical effects and actively tests competing hypotheses to resolve physical uncertainties. 
RoboAuto achieves superior performance across five benchmark families, attaining an 89.1\% success rate on short-horizon LIBERO-PRO tasks and more than twice the mean success rate of ASPIRE on LIBERO-PRO Long and LIBERO-Plus. It also outperforms ASPIRE and $\pi_{0.5}$ on three challenging long-horizon real-robot tasks requiring precise manipulation.
We further demonstrate that the hierarchical knowledge base transfers across simulators and embodiments, and that program search performance scales with the size of the knowledge base. Project page: \url{https://roboauto-ai.github.io/}.

\end{abstract}

\section{Introduction}
\label{sec:intro}

\begin{figure*}[t]
  \centering
  \includegraphics[width=\linewidth]{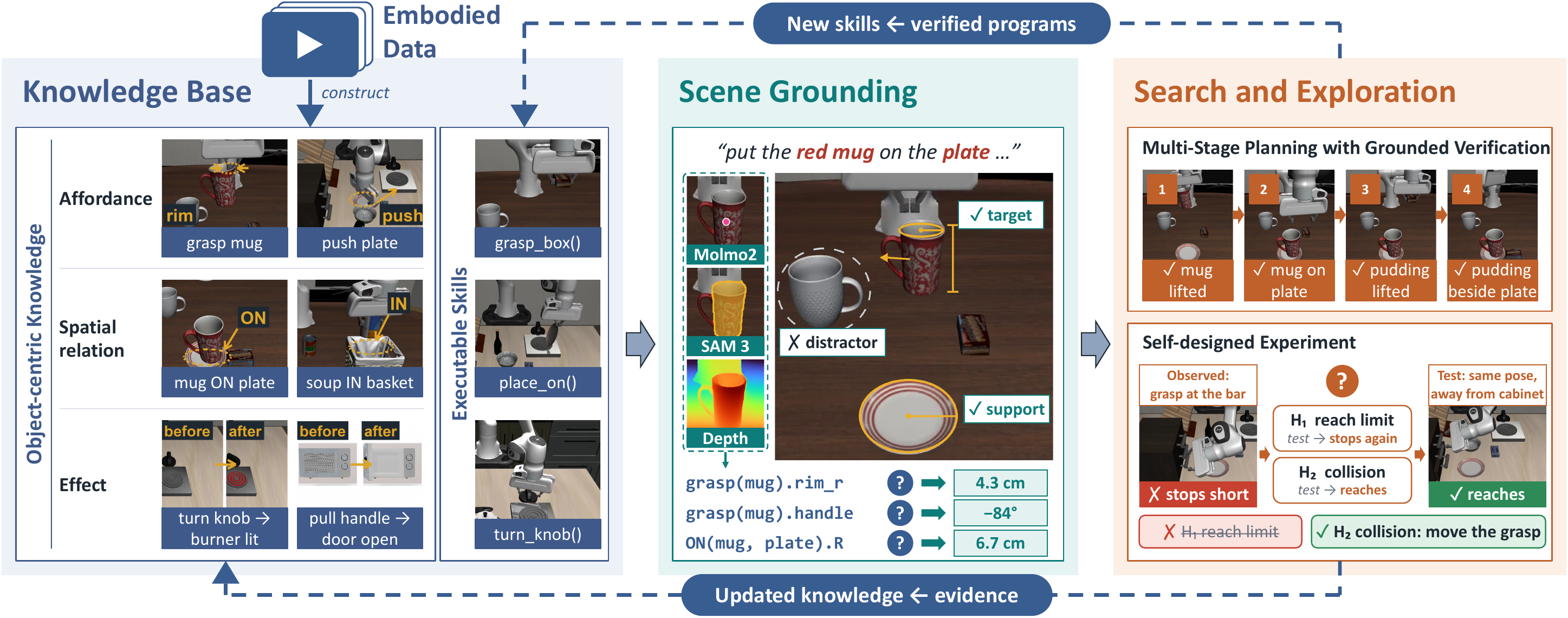}
  \caption{\textbf{Overview of RoboAuto.} The Hierarchical Knowledge Base (left) stores object-centric physical knowledge, comprising affordances, spatial relations, and action effects, together with executable skills. Physical Scene Grounding (middle) binds the entries relevant to a task to quantities measured in the current scene. Search and Exploration (right) plans the task as stages that end in verified state changes and designs experiments that separate competing physical hypotheses. Interaction evidence updates the knowledge, and verified programs return to the \kb as new skills. RoboAuto is evaluated across five benchmark families, LIBERO-PRO, LIBERO-PRO Long, LIBERO-Plus, Robosuite, and RoboTwin~2.0, where RoboTwin~2.0 tests transfer to a new simulator and embodiment.}
  \label{fig:teaser}
\end{figure*}

\begin{figure}[t]
  \centering
  \includegraphics[width=\linewidth]{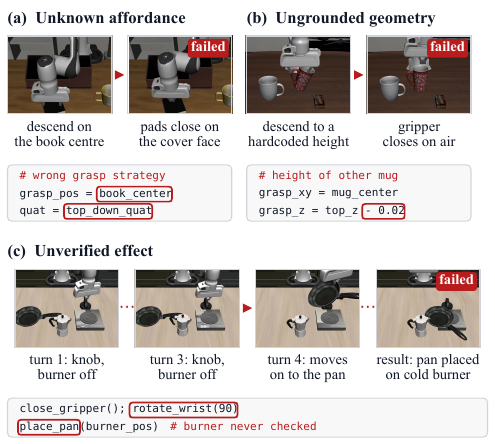}
  \caption{\textbf{Unresolved physical assumptions in our baseline runs.} In the illustrated ASPIRE and CaP-Agent0 runs, (a) an unsuitable contact strategy persists, (b) a grasp-height rule is transferred without grounding its geometric applicability, and (c) pan placement proceeds without verifying stove activation. The panels show the attempts, outcomes, and retained program choices; Appendix~\ref{sec:supp_behavior} provides the detailed case studies.}
  \label{fig:failures}
\end{figure}

Recent advances in autonomous research and coding agents built on large language models (LLMs)~\cite{yamada2025ai,novikov2025alphaevolve,zhang2026darwin} highlight the promise of a self-evolving problem-solving paradigm in which an agent generates an executable solution, evaluates its outcome, and uses the resulting evidence to improve subsequent attempts. Code-as-Policy~\cite{liang2023code} naturally extends this paradigm to robot manipulation by representing robot behaviors as executable programs. Given an instruction such as ``put the mug on the plate,'' a coding agent can generate a program over perception and control primitives, execute it in the environment, and revise it based on the resulting rollout.

Recent autonomous Code-as-Policy methods instantiate this generate, execute, and revise loop. CaP-Agent0 supports multi-turn program revision with structured feedback~\cite{fu2026cap}; ASPIRE exposes multimodal execution traces and accumulates validated repair knowledge in a skill library~\cite{lu2026aspire}; and RHO searches over policy repositories using reward and execution feedback~\cite{elmaaroufi2026rho}. These methods demonstrate the value of execution feedback and retained experience for robot program search.

However, \Cref{fig:failures} reveals persistent gaps in physical understanding in our baseline runs. In \Cref{fig:failures}(a), CaP-Agent0 repeatedly attempts a physically unsuitable top-down grasp on a book, indicating that repeated code-level search cannot always compensate for missing knowledge of object affordances. This type of failure motivates a hierarchical knowledge base that complements object-centric physical knowledge acquired from embodied data with executable skills accumulated through online task solving. In \Cref{fig:failures}(b), ASPIRE attempts to grasp a mug by descending to a hardcoded height transferred from another mug, causing the gripper to close on air. This failure shows that even a valid strategy must be instantiated using the geometry of the current scene, motivating a dedicated grounding stage that identifies task-relevant objects and physical quantities with visual tools. In \Cref{fig:failures}(c), CaP-Agent0 proceeds to place a pan despite failing to activate the burner, motivating multi-stage planning and grounded verification of intermediate effects. Moreover, observing a failed knob rotation does not establish how far the knob must turn. Resolving this uncertainty requires active physical exploration through autonomously designed experiments that test specific hypotheses.

Motivated by these observations, we introduce RoboAuto (\cref{fig:teaser}), a framework for autonomous robot program search through \emph{multi-level physical understanding}. Its hierarchical \kb combines object-centric physical knowledge about affordances, spatial relations, and action effects with executable skills. Physical knowledge is acquired from embodied data and refined through online experience, while the skill library expands as verified programs are discovered through task solving. For each task, RoboAuto retrieves relevant knowledge and skills and grounds them in the current scene, using visual tools, including Molmo2~\cite{clark2026molmo2} and SAM~3~\cite{carion2026sam}, to identify task-relevant objects and measure action-critical quantities. Through \emph{multi-stage planning with grounded verification}, it plans over stage dependencies and verifies each stage's physical outcome before proceeding. When available evidence leaves a physical uncertainty unresolved, RoboAuto formulates competing hypotheses and conducts \emph{self-designed experiments} to distinguish them. The resulting evidence refines the physical knowledge, while verified programs expand the skill library, improving both the current solution and future program search.

We evaluate RoboAuto on tasks from five benchmark families. On LIBERO-PRO Long~\cite{zhou2025libero}, RoboAuto achieves a mean success rate of 73.4\%, compared with 30.8\% for ASPIRE~\cite{lu2026aspire}, the strongest coding-agent baseline on average. On LIBERO-Plus~\cite{fei2026libero}, it reaches 90.3\%, compared with ASPIRE's 44.7\%, while achieving 89.1\% on the short-horizon LIBERO-PRO~\cite{zhou2025libero} tasks. On contact-rich Robosuite~\cite{zhu2020robosuite} tasks, RoboAuto obtains 90.6\% mean success, compared with ASPIRE's 81.0\%. To evaluate knowledge transfer, we apply the hierarchical knowledge base acquired in LIBERO~\cite{liu2023libero} to RoboTwin~2.0~\cite{chen2025robotwin}, which features a different simulator and a bimanual robot embodiment. RoboAuto achieves a 91.5\% success rate, compared with 24.5\% without a knowledge base. Knowledge-base scaling experiments further show that larger knowledge bases consistently improve program-search performance. RoboAuto also outperforms ASPIRE and $\pi_{0.5}$ on three challenging long-horizon real-robot tasks requiring precise manipulation.

Our contributions are:
\begin{itemize}
    \item We identify key limitations of existing autonomous Code-as-Policy methods in physical knowledge, scene grounding, intermediate verification, and active exploration.
    \item We introduce RoboAuto, which addresses these limitations through a hierarchical \kb, scene grounding, multi-stage planning with grounded verification, and autonomously designed experiments.
    \item Across five benchmark families, RoboAuto substantially improves program-search performance and demonstrates knowledge scaling and transfer across simulators and embodiments.
\end{itemize}

\section{Related Work}
\label{sec:related}

\paragraph{Robotic foundation models}
Learned visuomotor policies provide increasingly general foundations for robot control. Diffusion-based policies model multimodal action distributions and have been extended to improve spatial generalization, inference efficiency, and closed-loop responsiveness~\cite{chi2025diffusion,ze20243d,prasad2024consistency,gao2024manicm,chen2025responsive}. At a larger scale, vision-language-action models learn generalist language-conditioned control from diverse tasks, environments, and robot embodiments~\cite{zitkovich2023rt,team2024octo,kim2024openvla,black2024pi_0,intelligence2025pi_,bjorck2025gr00t,team2025gemini,team2025gemini15,zhang2026prts}. World-action models complement this paradigm by coupling action generation with predictions of future visual or latent world states~\cite{zhu2025unified,hu2024video,ye2026world,yuan2026fast,zhang2026imagewam,zhang2026v}. Whereas these robotic foundation models acquire behavior primarily by training parameterized action policies on large-scale data, RoboAuto introduces a fundamentally different paradigm in which explicit robot programs are autonomously searched and refined through grounded physical understanding, execution-based verification, and active exploration.

\paragraph{Code-as-Policy and autonomous robot self-improvement}
Code as Policies~\cite{liang2023code} represents robot behavior as programs that compose perception and control primitives, making generated behavior directly inspectable and revisable. CaP-X~\cite{fu2026cap} studies how primitive design, perceptual grounding, and interaction shape this interface, while CaP-Agent0 extends it to multi-turn program revision through execution feedback. More broadly, autonomous research agents~\cite{yamada2025ai,novikov2025alphaevolve} and recursive self-improvement systems~\cite{zhang2026darwin,wang2023voyager} seek to turn experience from individual executions into persistent capabilities. In robotics, RHO~\cite{elmaaroufi2026rho} searches over policy repositories, ASPIRE~\cite{lu2026aspire} distills validated repairs into reusable skills, RATs~\cite{zhang2026playful} acquires skills through self-directed play, and RoboRSI~\cite{wen2026roborsi} organizes reusable capabilities through skill contracts and validated refinement. OpenETA~\cite{chen2026eta} and ENPIRE~\cite{xiao2026enpire} similarly improve retained experience or learned policies through physical interaction. Recent work further formalizes physical recursive self-improvement (RSI) and explores recursive improvement through real-to-sim-to-real policy adaptation and simulation-guided refinement of prompts and symbolic skills~\cite{lang2026embodirsi,zhou2026improves,wang2026reconstruct}. Beyond improving policies, prompts, or skill code, RoboAuto makes physical understanding itself a persistent and continually improvable substrate for program search by integrating a hierarchical knowledge base, scene grounding, multi-stage grounded verification, and hypothesis-driven active exploration.

\paragraph{Physical knowledge and scene grounding}
Knowledge-based robotics connects reusable object and action knowledge with perception, planning, and execution~\cite{tenorth2013knowrob,beetz2018know,ye2026unidomain,ye2026uniplan,chen2024language}. MEMORA~\cite{yu2026memora} further transforms egocentric experience into structured, editable memory for grounded robot planning. In parallel, scene-grounding methods construct general-purpose 3D representations~\cite{jatavallabhula2023conceptfusion,gu2024conceptgraphs} or extract task-specific spatial constraints and geometric information~\cite{huang2023voxposer,huang2024rekep,sun2025executable}. RoboAuto builds on these directions by maintaining reusable object-centric knowledge and selectively grounding task-relevant objects and action-critical quantities during program search, while using execution evidence and autonomously designed experiments to refine its knowledge online.

\section{Method}
\label{sec:method}

RoboAuto enables autonomous robot program search through multi-level physical understanding. It consists of three components (\cref{fig:teaser}). The \textbf{Hierarchical Knowledge Base} (\cref{sec:kb}) maintains object-centric physical knowledge learned from embodied data and executable skills acquired through online task solving. \textbf{Physical Scene Grounding} (\cref{sec:ground}) identifies task-relevant objects and measures action-critical quantities to instantiate candidate programs in the current scene. \textbf{Search and Exploration} (\cref{sec:search}) decomposes long-horizon tasks, verifies intermediate outcomes, and designs targeted experiments to resolve physical uncertainties.

\subsection{Hierarchical Knowledge Base}
\label{sec:kb}

Robot program search benefits from reusable experience at two different levels of abstraction. General physical knowledge describes what interactions are appropriate and what effects they should produce, while executable skills encode how those interactions are implemented with perception and control primitives. RoboAuto separates these two forms of experience in a hierarchical knowledge base. Its high-level layer contains object-centric physical knowledge that transfers across tasks and scenes, while its low-level layer contains executable skills that can be directly reused and composed during program search. This separation allows physical knowledge to guide the selection, grounding, and verification of robot programs without tying it to a particular implementation.

\paragraph{Object-centric physical knowledge} Although online task solving can continually expand an agent’s experience, discovering appropriate interaction strategies solely through program search can be difficult. Without object-centric physical knowledge, the agent may continue refining an unsuitable strategy instead of recognizing that a fundamentally different interaction is required. RoboAuto therefore organizes reusable physical knowledge into entries indexed by object class and constructs these entries from both embodied demonstration data and online task-solving experience. Each entry captures three complementary forms of knowledge. Affordances describe suitable interaction strategies for an object; spatial relations characterize task-relevant configurations such as support, containment, and relative position; and action effects and preconditions specify the state changes produced by an action and the conditions required for those changes. RoboAuto constructs these entries from embodied data by analyzing demonstrated interactions, object configurations, and state transitions, and continually refines them using evidence collected during online task solving. To make this knowledge applicable across scenes, each entry also specifies the conditions under which a strategy is applicable and the scene-dependent quantities required to instantiate it. For each quantity, the entry records its purpose and how it can be measured, while leaving its value to be established in the current scene. Physical Scene Grounding (\cref{sec:ground}) measures these quantities for the current objects and robot, allowing the same object-centric knowledge to guide program search across different scenes.

\paragraph{Executable skills} The second layer stores verified robot programs as executable skills. Whereas object-centric knowledge captures transferable physical regularities, executable skills implement concrete behaviors using available perception and control primitives. RoboAuto retrieves these skills as candidate implementations or building blocks during program search, reducing the need to rediscover previously successful behaviors. Programs verified through online task solving are added to the skill library, allowing the executable layer to expand continually. Appendix~\ref{sec:supp_method_kb} provides more details.

\subsection{Physical Scene Grounding}
\label{sec:ground}

Object-centric knowledge describes reusable interaction strategies and their grounding requirements, but its scene-dependent quantities must be established for each task. Given an instruction and the current observation, RoboAuto identifies the task-relevant objects, assigns their roles in the intended interaction, and associates them with the retrieved knowledge entries and executable skills. It then follows the entries’ measurement procedures to check strategy applicability and obtain the required quantities. Rather than constructing an exhaustive representation of the entire scene, this process is scoped by the task and the intended interaction. RoboAuto combines visual perception models, including Molmo2~\cite{clark2026molmo2} and SAM~3~\cite{carion2026sam}, with depth-camera observations to identify and segment relevant objects and obtain action-critical information such as grasp locations, metric dimensions, orientations, and spatial relations. Grounding can also involve active viewpoint selection. We introduce \textbf{active viewpoint selection}, which enables the agent to move the robot arm to obtain a more informative camera view when the initial observation is insufficient, thereby improving the accuracy of grounding task-relevant quantities. The resulting grounded representation binds general physical knowledge to concrete objects and measurements, providing the coding agent with the information needed to select a feasible strategy and instantiate it as an executable program. 
Scene grounding also records which physical conditions have been established and which remain uncertain. Grounded values inform program generation and planning, while unresolved conditions are passed to Search and Exploration (\cref{sec:search}) for further investigation. Grounding remains open to revision throughout program search. An execution outcome may invalidate an earlier measurement, reveal a previously unmodeled condition, or change the scene itself. RoboAuto can then remeasure relevant quantities and reassess knowledge applicability to guide the next candidate program. In this way, scene grounding provides a continually updated connection between reusable physical knowledge and the physical state in which the robot acts. Appendix~\ref{sec:supp_method_ground} provides more details.

\subsection{Search and Exploration}
\label{sec:search}

\paragraph{Multi-stage planning with grounded verification} Long-horizon tasks require prerequisite physical changes to be verified before subsequent actions proceed. RoboAuto decomposes each task into stages, each defined by an intended change in object state. Physical dependencies determine their order, while expected effects define the verification criteria. RoboAuto then searches sequentially over the stages. For the current stage, it generates and executes a candidate program together with an executable verifier that evaluates the resulting state from a fresh observation. Search proceeds to the next stage only after the current stage has been verified; otherwise, subsequent execution is stopped and revision remains focused on the incomplete stage. This procedure prevents failures from propagating to later stages, localizes search to a specific physical transition, and preserves the verified program prefix instead of repeatedly regenerating the entire long-horizon solution.

Because the executable verifier may misinterpret an observation, RoboAuto uses an independent auditor to review unsuccessful task executions. Given the stage plan, relevant physical knowledge, execution logs, and synchronized scene and wrist-camera videos, the auditor determines whether each stage achieved its intended object state, and its judgments replace the corresponding verifier results. The coding agent uses this feedback to determine whether the stage program, executable verifier, or overall plan requires revision. When the evidence indicates that a stage is incorrectly defined, ordered under the wrong dependency, or physically infeasible as planned, RoboAuto revises the stage decomposition and regenerates the plan before resuming sequential search, reopening completed stages when their outcomes must be re-established.

\paragraph{Self-designed experiments} Existing autonomous Code-as-Policy methods primarily acquire evidence by executing candidate robot programs and revising them afterward. Because these executions are intended to complete the task rather than investigate a specific uncertainty, their evidence may not isolate the physical assumption responsible for a failure. RoboAuto instead actively explores the scene through autonomously designed experiments. Based on its current physical understanding and scene observations, it formulates physical hypotheses, predicts their observable consequences, and writes dedicated experiment code, separate from the candidate robot program, that specifies the physical intervention and the observations required to evaluate the hypotheses. For example, RoboAuto can test several candidate grasp poses through controlled grasp-and-lift trials and incorporate the successful pose into the robot program without repeatedly executing the full task. The outcome can support, reject, or refine a hypothesis, providing targeted evidence for subsequent program search. It also updates the corresponding object-centric knowledge entry, so that exploration during the current task improves future task solving. \Cref{sec:supp_method_search} provides more details, and \cref{sec:supp_prompt_listings} reproduces the prompts.

\section{Experiments}
\label{sec:experiments}

\begin{figure*}[t]
  \centering
  \includegraphics[width=\linewidth]{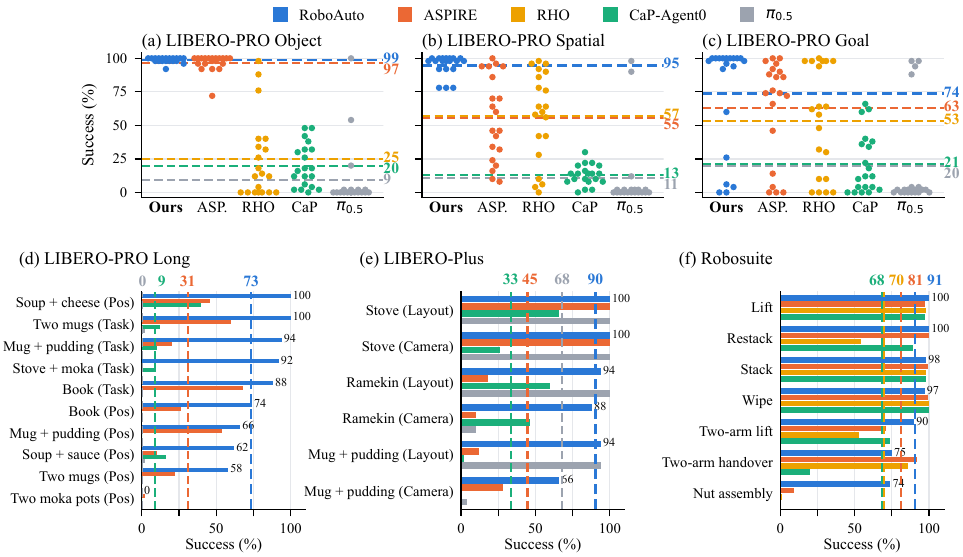}
  \caption{\textbf{Program success across benchmarks.} (a--c) Each dot is a LIBERO-PRO task under position or task perturbation. (d--f) Horizontal bars give each method's success on every LIBERO-PRO Long, LIBERO-Plus, and Robosuite setting, ordered from the top by RoboAuto's success, which is printed beside its bar. Dashed lines and their labels give method means.}
  \label{fig:main}
\end{figure*}

\subsection{Experimental Setup}
\label{sec:setup}

\paragraph{Benchmarks and evaluation protocol}
We evaluate RoboAuto across five benchmark families spanning program robustness, long-horizon manipulation, contact-rich control, and cross-embodiment knowledge transfer. LIBERO-PRO~\cite{zhou2025libero} evaluates short-horizon Object, Goal, and Spatial tasks under position and task perturbations, while LIBERO-PRO Long focuses on multi-stage goals. LIBERO-Plus~\cite{fei2026libero} introduces camera and scene-layout variations. Robosuite~\cite{zhu2020robosuite} includes contact-rich single- and dual-arm manipulation tasks, and RoboTwin~2.0~\cite{chen2025robotwin} evaluates knowledge transfer to a different simulator and bimanual embodiment. Program search and final evaluation use disjoint scenes, testing whether the discovered programs generalize beyond the scenes in which they were developed.

\paragraph{Baselines and implementation}
We compare RoboAuto with three autonomous Code-as-Policy methods and a generalist vision-language-action (VLA) model. CaP-Agent0~\cite{fu2026cap} iteratively revises programs using visual feedback and reuses automatically synthesized skills. ASPIRE~\cite{lu2026aspire}, our primary baseline, conducts evolutionary program search and retains skills from prior repairs. RHO~\cite{elmaaroufi2026rho} improves policy repositories through reward and execution feedback, while the official LIBERO-finetuned $\pi_{0.5}$~\cite{intelligence2025pi_} checkpoint serves as the generalist VLA baseline in simulation. All evaluations build on CaP-Bench~\cite{fu2026cap}, which we extend to all five benchmark families, and all autonomous Code-as-Policy methods write programs over the same set of perception and control primitives, including depth-camera observations, Molmo2~\cite{clark2026molmo2} pointing, SAM~3~\cite{carion2026sam} segmentation, and Contact-GraspNet~\cite{sundermeyer2021contact} grasp proposals. RoboAuto and all reproduced Code-as-Policy baselines use Claude Opus~4.6 with a 1M-token context window as their coding agent. The initial object-centric knowledge base is built from demonstration videos from the original LIBERO-90 benchmark, disjoint from all evaluation tasks. Further implementation and evaluation details appear in \cref{sec:supp_setup}.

\subsection{Main Results}
\label{sec:main}

\Cref{fig:main} shows that RoboAuto consistently achieves the highest mean success across all evaluated benchmark suites, substantially outperforming the Code-as-Policy and VLA baselines under a wide range of manipulation settings. This advantage also holds in every category when all evaluation settings are grouped by whether their background and objects match those of the LIBERO-90 demonstrations (\cref{sec:supp_demo_overlap}).

\paragraph{LIBERO-PRO and LIBERO-Plus: generalization across tasks and scenes}
LIBERO-PRO evaluates generalization under position and task perturbations across its Object, Spatial, and Goal suites, while LIBERO-Plus introduces camera and scene-layout changes. On LIBERO-PRO Object, RoboAuto and ASPIRE both perform near ceiling at 98.7\% and 96.6\%, respectively. On Spatial, which requires identifying the target object through its position relative to other objects and placing it at a specified location, RoboAuto achieves 94.8\% success compared with 55.3\% for ASPIRE. On Goal, which spans diverse interactions such as opening drawers, turning on the stove, and pushing objects, it achieves 73.7\% compared with 63.1\%. On LIBERO-Plus, RoboAuto achieves 90.3\% success compared with 44.7\% for ASPIRE. These results show that RoboAuto can retain reusable interaction knowledge while grounding relevant objects and action-critical quantities for new tasks, layouts, and observations.

\paragraph{LIBERO-PRO Long: multi-stage manipulation}
LIBERO-PRO Long tests whether a method can manage physical dependencies across a long-horizon task. RoboAuto reaches 73.4\% mean success, more than doubling ASPIRE's 30.8\%. RoboAuto decomposes each task into stages, searches over them sequentially, and verifies each intermediate outcome before proceeding, preventing unresolved failures from propagating to later stages.

\paragraph{Robosuite: cross-benchmark transfer to contact and coordination}
Robosuite tests precise contact, geometric alignment, and inter-arm coordination in settings distinct from the LIBERO tasks used to initialize RoboAuto. No Robosuite demonstrations are used to construct the knowledge base. RoboAuto nevertheless achieves 90.6\% mean success versus 81.0\% for ASPIRE, including 74\% on nut assembly while every baseline remains below 10\%. These results show that object-centric physical knowledge acquired from embodied data can transfer across benchmarks and be re-grounded for contact-sensitive manipulation in substantially different settings.

\paragraph{Generality and transferability of object-centric knowledge}
The initial object-centric knowledge is constructed from LIBERO-90 data and is designed to transfer: its entries describe qualitative strategies and the quantities to be measured in each scene, contain no control commands, physical values, task instructions, or specific object instances, and are organized by general affordance, spatial, and effect categories rather than by task or object (\cref{sec:supp_demo_overlap}). Grouping evaluation settings by similarity to LIBERO-90, RoboAuto achieves the highest mean success in every group, including settings with both unseen scene layouts and unseen objects (84.3\% vs.\ 68.8\%; \cref{tab:demo_overlap}). Adding the knowledge base to ASPIRE improves it on several tasks but does not close the gap to RoboAuto (\cref{sec:kb_aspire}), indicating that RoboAuto's gains also rely on scene grounding and grounded multi-stage planning.

\subsection{Further Experiments and Analysis}
\label{sec:ablation}

This section aims to study the following questions: 

\begin{itemize}
    \item \textbf{RQ1: How does each component of RoboAuto contribute to robot program search?} (\cref{sec:ablation_study,sec:supp_ablation})
    \item \textbf{RQ2: How transferable is the hierarchical knowledge base across simulators and robot embodiments?} (\cref{sec:generalize,sec:supp_transfer})
    \item \textbf{RQ3: How does scaling the number of available hierarchical knowledge-base entries affect program search?} (\cref{sec:scaling,sec:supp_kb_search})
    \item \textbf{RQ4: How does language model capability affect RoboAuto's performance?} (\cref{sec:supp_opus5})
    \item \textbf{RQ5: How sensitive is RoboAuto to randomness across independent program-search runs?} (\cref{sec:supp_variance})
    \item \textbf{RQ6: How does RoboAuto's token usage compare with existing program-search methods?} (\cref{sec:supp_tokens})
\end{itemize}

\begin{figure}[t]
  \centering
  \includegraphics[width=\linewidth]{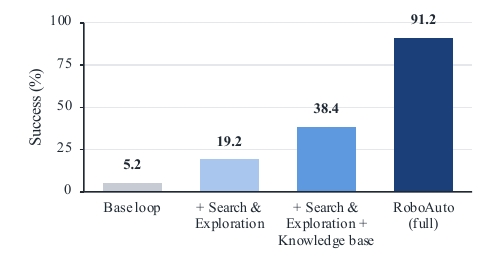}
  \caption{\textbf{Ablations of the three components.} Mean final-program success over five tasks as Search and Exploration, the Hierarchical Knowledge Base, and Physical Scene Grounding are cumulatively added to the base loop.}
  \label{fig:ablation}
\end{figure}

\subsubsection{Ablation Studies}
\label{sec:ablation_study}

The base loop writes a task program, runs it, and revises the whole program based on the execution results. \Cref{fig:ablation} reports mean final-program success over five tasks as the three components are added cumulatively. The base loop makes little progress and reaches only 5.2\%. Adding Search and Exploration raises success to 19.2\%, supporting stage-wise verification and targeted experiments as a means of turning execution feedback into useful revisions. Adding the Hierarchical Knowledge Base further raises success to 38.4\%, indicating that reusable interaction strategies help search progress beyond what structured revision alone achieves. However, this variant still falls well below the full RoboAuto system. The full RoboAuto system adds Physical Scene Grounding, producing the largest gain and reaching 91.2\%. This gap highlights the need to ground stored knowledge in the current objects and geometry. Per-task results and case studies appear in \cref{sec:supp_ablation,sec:supp_component_overview}.

\subsubsection{Hierarchical Knowledge Base Transferability}
\label{sec:generalize}

\begin{figure}[t]
  \centering
  \includegraphics[width=\linewidth]{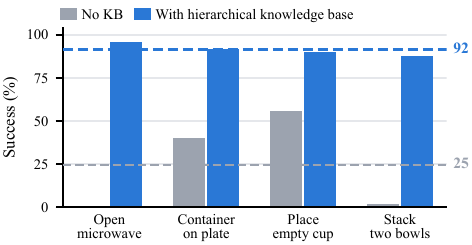}
  \caption{\textbf{Hierarchical knowledge base transfer to RoboTwin~2.0.} Success on four RoboTwin~2.0 tasks without retained knowledge and with the hierarchical knowledge base learned from LIBERO. Dashed lines indicate mean success.}
  \label{fig:robotwin}
\end{figure}

The Robosuite results in \cref{fig:main} demonstrate the cross-benchmark transfer of object-centric knowledge learned from LIBERO. RoboTwin~2.0~\cite{chen2025robotwin} provides a more stringent test by changing the simulator, robot embodiment, layouts, tasks, and control primitives. Success increases from 24.5\% without retained knowledge to 91.5\% with the hierarchical knowledge base (\cref{fig:robotwin}), demonstrating its substantial benefit for program search under cross-simulator and cross-embodiment shifts. Per-task success and token usage appear in \cref{sec:supp_transfer,sec:supp_tokens}.

\begin{figure}[t]
  \centering
  \includegraphics[width=\linewidth]{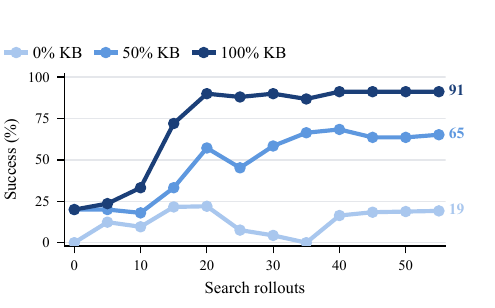}
  \caption{\textbf{Knowledge availability and program search.} Mean test success versus search rollouts on the five ablation tasks with 0\%, 50\%, and 100\% of the \kb. The 0\% and 100\% conditions are ``+ Search \& Exploration'' and RoboAuto in \cref{fig:ablation}.}
  \label{fig:scaling}
\end{figure}

\subsubsection{Knowledge Scaling}
\label{sec:scaling}

We study how the availability of hierarchical \kb entries affects program search by comparing  0\%, randomly sampled 50\%, and 100\% knowledge availability on five tasks under the same rollout budget (\cref{fig:scaling}). The half and full \kb conditions both begin at 20\% mean success but diverge as search proceeds. After 20 rollouts, the full \kb condition reaches 90.0\%, compared with 57.2\% for the half \kb condition. Because both conditions start from the same success, the gap reflects how additional entries help the agent interpret execution outcomes and choose subsequent revisions. With the full \kb, every task first reaches 80\% success within 19 rollouts, whereas over the full budget only three tasks reach this level with half of the \kb entries and one task without the \kb. Scaling knowledge availability therefore improves both search efficiency and the range of interactions solved within a fixed rollout budget. Per-task results appear in \cref{sec:supp_kb_search}.

\begin{figure}[t]
  \centering
  \includegraphics[width=\linewidth]{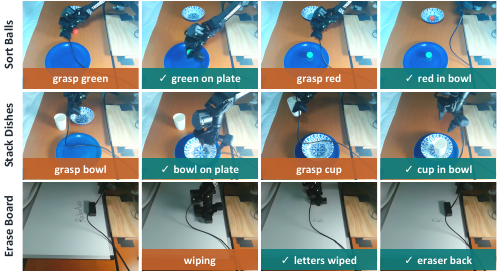}
  \caption{\textbf{Real-robot rollouts.} RoboAuto programs on a WidowX AI arm for three tasks. Each row shows four chronological scene-camera keyframes.}
  \label{fig:real_robot}
\end{figure}

\subsection{Real-Robot Experiments}
\label{sec:real_robot}

\begin{table}[!htbp]
  \centering
  \caption{\textbf{Real-robot evaluation.} Mean stage-completion score over 20 trials per task. \Cref{fig:real_robot} shows example rollouts.}
  \label{tab:real_robot}
  \small
  \setlength{\tabcolsep}{4pt}
  \renewcommand{\arraystretch}{1.12}
  \begin{tabular}{@{}lccc@{}}
    \toprule
    Task & RoboAuto & ASPIRE & $\pi_{0.5}$ \\
    \midrule
    Sort Balls & 100.00\% & 30.00\% & 0.00\% \\
    Stack Dishes & 100.00\% & 3.75\% & 0.00\% \\
    Erase Board &  95.83\% & 0.00\% & 0.00\% \\
    \bottomrule
  \end{tabular}
\end{table}

We further compare RoboAuto with ASPIRE and the DROID-finetuned $\pi_{0.5}$ checkpoint on three long-horizon, precision-manipulation tasks with a WidowX AI arm. RoboAuto reuses its hierarchical knowledge base from LIBERO, ASPIRE reuses its LIBERO-derived skill library, and both use Opus~5 with a 1M-token context window for program search. \textbf{Sort Balls} places two slippery ping-pong balls in their receptacles, \textbf{Stack Dishes} stacks a low-rimmed bowl and a cup, and \textbf{Erase Board} erases ``RoboAuto'' with a compliant eraser and returns it to its original position. \Cref{sec:supp_real_robot} describes the hardware, tasks, and evaluation protocol. As shown in \cref{tab:real_robot}, RoboAuto achieves near-perfect scores on all three tasks, whereas ASPIRE makes limited progress and $\pi_{0.5}$ makes none, showing that the hierarchical knowledge base learned in simulation transfers to a real robot and unseen tasks.

\section{Conclusion}
\label{sec:conclusion}

We introduced RoboAuto, a framework for autonomous robot program search that combines a hierarchical knowledge base, scene grounding, stage-wise verified search, and self-designed experiments. Across five benchmark families, RoboAuto consistently improves task success and more than doubles ASPIRE's performance on LIBERO-PRO Long and LIBERO-Plus. Knowledge acquired in LIBERO also transfers to a new simulator and bimanual embodiment in RoboTwin~2.0, reaching 91.5\% success versus 24.5\% without it. These results show that transferable physical knowledge and executable robot programs play complementary roles.

\paragraph{Limitations}
RoboAuto searches autonomously only in simulation. On real robots, a human operator still resets the scene after each rollout and keeps the arm and its surroundings safe. Automating scene resets and safety monitoring would bring the same autonomy to real-world deployment.

\clearpage

{
    \small
    \bibliographystyle{ieeenat_fullname}
    \bibliography{main}
}

\clearpage

\appendix
\crefalias{section}{appendix}

\section{Detailed Experimental Setup}
\label{sec:supp_setup}

This section specifies the benchmark tasks, RoboAuto search and evaluation protocol, system configuration, and baseline implementations. Quantitative results are reported separately in the main paper and subsequent appendix sections.

\subsection{Simulation Benchmarks}

\subsubsection{LIBERO-PRO}
\label{sec:supp_libero_pro}

LIBERO-PRO~\cite{zhou2025libero} evaluates adaptation to controlled task and scene changes. We use its Object, Spatial, and Goal suites. Object requires picking a specified object among distractors and placing it in a basket. Spatial requires identifying the target object through its position relative to other objects and placing it at a specified location. Goal spans diverse interactions such as opening drawers, turning on the stove, and pushing objects within a shared scene. For each suite, we separately evaluate position perturbations, which relocate task-relevant objects, and task perturbations, which modify the goal condition and instruction. Each suite contains ten tasks under each perturbation, producing 60 settings in total. \Cref{tab:libero_pro} reports the suite-level evaluation, while Tables~\ref{tab:libero_pro_object}--\ref{tab:libero_pro_spatial} provide the complete task-level matrix.

\subsubsection{LIBERO-PRO Long}
\label{sec:supp_long}

LIBERO-PRO Long evaluates programs that must produce several dependent physical state changes in the correct order. We randomly select ten task--perturbation settings from the Long suite, covering both task and position perturbations. These tasks require program search to maintain physical progress over an extended horizon and avoid executing later actions before their prerequisites have been established. All methods are evaluated on the same ten settings listed in \cref{tab:long}.

\subsubsection{LIBERO-Plus}
\label{sec:supp_libero_plus}

LIBERO-Plus~\cite{fei2026libero} evaluates robustness to visual and geometric changes that preserve the underlying manipulation objective. We randomly select three base tasks from LIBERO-Plus. Each task is evaluated under one camera variation and one layout variation, yielding six settings. For each task and variation type, we use the highest available difficulty level: level~5 for all settings except the camera variation of the stove task, for which the highest available level is~4. Camera variations change the workspace viewpoint and framing, while layout variations change object positions and spatial configurations. These changes require the agent to re-identify task-relevant objects and re-estimate scene-dependent quantities rather than reuse parameters tuned to the original scene.

\subsubsection{Robosuite}
\label{sec:supp_robosuite}

From Robosuite~\cite{zhu2020robosuite}, we use seven single- and dual-arm manipulation tasks. Lift evaluates grasping and transport; Stack and Restack require precise relative placement; Wipe requires sustained surface contact; and Nut Assembly requires accurate geometric alignment. Two-Arm Lift and Two-Arm Handover additionally test coordination between manipulators. Although LIBERO and Robosuite both use MuJoCo, these tasks introduce different interaction structures, contact requirements, and robot configurations. The complete task set is listed in \cref{tab:robosuite}.

\subsubsection{RoboTwin~2.0}
\label{sec:supp_robotwin}

RoboTwin~2.0~\cite{chen2025robotwin} evaluates transfer to a different simulator and robot embodiment. It uses a bimanual robot in a SAPIEN/PhysX environment~\cite{xiang2020sapien}, changing the gripper geometry, control primitives, contact dynamics, layouts, and task definitions. To evaluate the transfer of the knowledge base, we compare search without retained knowledge and with the hierarchical knowledge base learned from LIBERO under the same search protocol. \Cref{tab:supp_robotwin} lists the four evaluated tasks and both conditions.

\subsection{RoboAuto Evaluation Protocol}
\label{sec:supp_protocol}

\paragraph{Search and test splits}
Unless otherwise specified in \cref{tab:supp_protocol}, program search uses 15 training scenes with seeds 110000--110014, and final-program evaluation uses 50 test scenes with seeds 1--50. Robosuite uses 100 test scenes with seeds 1--100. The two sets are disjoint. Program search uses only training-scene evidence; test-scene outcomes are never exposed to the agent or used to update either layer of the knowledge base. After search terminates, the last program is fixed and executed independently on every test scene. The benchmark environment predicate determines task-level success, and suite means weight all listed task settings equally.

\paragraph{Rollout accounting and termination}
One rollout is defined as executing a single program in a single training scene. Search runs over a growing subset of the training scenes, the search pool: each candidate is first run on two scenes, and the pool grows from five to ten and then fifteen scenes as each subset is solved. Search terminates when all training scenes are solved or the rollout budget in \cref{tab:supp_protocol} is exhausted. All program executions for active viewpoint selection during scene grounding and for self-designed experiments count toward the same rollout budget. The execution and revision protocols are detailed in \cref{sec:supp_prompts}.

Final-program evaluation follows the protocol of CaP-Bench and ASPIRE, with 50 test scenes per LIBERO task and 100 per Robosuite task, and all reproduced baselines use the same training and test scenes.

\begin{table*}[!tbp]
  \centering
  \begin{threeparttable}
  \caption{\textbf{Scene seeds and rollout budgets.} Seed intervals include both endpoints. The number of settings counts task--perturbation combinations.}
  \label{tab:supp_protocol}
  \small
  \setlength{\tabcolsep}{4pt}
  \renewcommand{\arraystretch}{1.12}
  \begin{tabular}{@{}lcrrc@{}}
    \toprule
    Suite & Settings & Training seeds & Test seeds & Rollouts \\
    \midrule
    LIBERO-PRO Object & 20 & 110000--110014 & 1--50 & 45 \\
    LIBERO-PRO Goal & 20 & 110000--110014 & 1--50 & 45 \\
    LIBERO-PRO Spatial & 20 & 110000--110014 & 1--50 & 45 \\
    LIBERO-PRO Long & 10 & 110000--110014 & 1--50 & 55 \\
    LIBERO-Plus Spatial & 2 & 110000--110014 & 1--50 & 45 \\
    LIBERO-Plus Goal & 2 & 110000--110014 & 1--50 & 45 \\
    LIBERO-Plus Long-derived & 2 & 110000--110014 & 1--50 & 55 \\
    Robosuite & 7 & 110000--110014 & 1--100 & 45 \\
    RoboTwin~2.0 & 4 & 110000--110014 & 1--50 & 11 \\
    \bottomrule
  \end{tabular}
  \begin{tablenotes}[flushleft]\footnotesize
  \item[] Program search uses 15 training scenes; final-program evaluation uses 50 test scenes, except Robosuite, which uses 100. The two LIBERO-Plus Long-derived settings are mug-and-pudding tasks under camera and layout perturbations. Ablations use the budget of the corresponding task.
  \end{tablenotes}
  \end{threeparttable}
\end{table*}

\subsection{RoboAuto Configuration}
\label{sec:supp_roboauto_config}

\paragraph{Code-as-Policy interface}
We extend the CaP-Bench evaluation interface~\cite{fu2026cap} to support all five benchmark families. Benchmark-specific wrappers expose perception and control primitives to generated programs and return execution logs, camera observations, and the environment success verdict. RoboAuto writes staged task programs over these primitives. Self-designed experiments use separate experiment and control files but execute through the same benchmark interface.

\paragraph{Initial knowledge construction}
The initial object-centric knowledge base is constructed from successful LIBERO demonstration videos, using exactly one video per source task in LIBERO-90. \Cref{sec:supp_demo_overlap} groups all evaluation settings by their relation to these demonstrations.

\paragraph{Online knowledge and skill updates}
All evaluation tasks begin from the knowledge condition specified for that experiment. Object-centric revisions and executable skills are created only from evidence collected during program search. Before each independently evaluated task, RoboAuto is restored to the corresponding initial condition. Knowledge-scaling experiments vary only the availability of stored entries while keeping the search protocol fixed.

\paragraph{Coding and perception models}
RoboAuto uses Claude Opus~4.6 with a 1M-token context window as its coding agent, referred to below as the coder, for program generation and revision. Physical Scene Grounding combines Molmo2~\cite{clark2026molmo2} and SAM~3~\cite{carion2026sam} with depth-camera observations to identify task-relevant objects and estimate action-critical quantities. The prompts, knowledge representation, and grounding procedures are described in \cref{sec:supp_prompt_listings,sec:supp_method_kb,sec:supp_method_ground}.

\subsection{Baselines and Comparison Protocol}
\label{sec:supp_baselines}

We compare RoboAuto with three autonomous Code-as-Policy methods and one generalist vision-language-action model. These baselines cover iterative program revision, accumulated executable experience, repository-level policy search, and direct action generation. All autonomous Code-as-Policy methods write programs over the same set of perception and control primitives, including depth-camera observations, Molmo2~\cite{clark2026molmo2} pointing, SAM~3~\cite{carion2026sam} segmentation, and Contact-GraspNet~\cite{sundermeyer2021contact} grasp proposals.

\paragraph{CaP-Agent0}
CaP-Agent0~\cite{fu2026cap} generates robot programs over perception and control primitives and revises them across multiple turns using execution logs and visual feedback. It can retain automatically synthesized skills from previous task-solving experience. We compare against its full configuration, which adds a synthesized skill library and parallel reasoning to the multi-turn visual-differencing tier (M4) of CaP-Bench. CaP-Agent0 has no separate search phase and uses no training scenes. Instead, it adapts at test time within each evaluation episode: it executes an initial program on the test scene and may revise it up to ten times based on execution feedback, without resetting the environment, so each evaluation episode allows up to eleven program executions. In our reproduction on LIBERO-PRO Long and LIBERO-Plus, we keep its released configuration unchanged and use Claude Opus~4.6 with a 1M-token context window for all of its model queries.

\paragraph{ASPIRE}
ASPIRE~\cite{lu2026aspire} performs evolutionary search over candidate robot programs using multimodal execution traces and retains a skill library from repair experience.

\paragraph{RHO}
RHO~\cite{elmaaroufi2026rho} searches over a repository of robot policies and improves the repository using reward and execution feedback.

\paragraph{\texorpdfstring{$\boldsymbol{\pi}_{0.5}$}{pi0.5}}
$\pi_{0.5}$~\cite{intelligence2025pi_} is a generalist vision-language-action model that directly predicts robot actions from visual observations and language instructions, providing a non-programmatic comparison with a pretrained visuomotor policy.

\paragraph{Reported and reproduced baseline results}
On LIBERO-PRO Object, Goal, and Spatial and on Robosuite, the results of CaP-Agent0, ASPIRE, and RHO are from the published results in CaP-X~\cite{fu2026cap}, ASPIRE~\cite{lu2026aspire}, and RHO~\cite{elmaaroufi2026rho}, respectively. On LIBERO-PRO Long and LIBERO-Plus, we reproduce ASPIRE and CaP-Agent0 from their released codebases using their original experimental configurations without modification. To control for coding-model capability, both use Claude Opus~4.6 with a 1M-token context window, matching RoboAuto. Complete execution records from these runs are retained for the token-usage comparison. RHO has not released the code for its robot experiments, so it cannot be reproduced on LIBERO-PRO Long and LIBERO-Plus. The $\pi_{0.5}$ results on LIBERO-PRO and LIBERO-PRO Long are from the published results in LIBERO-PRO~\cite{zhou2025libero}, which evaluates the official LIBERO-finetuned checkpoint. On LIBERO-Plus, we evaluate the same published LIBERO-finetuned $\pi_{0.5}$ checkpoint ourselves on the same 50 test scenes. We do not report $\pi_{0.5}$ on Robosuite because no $\pi_{0.5}$ checkpoint has been released for Robosuite, and the released LIBERO and DROID checkpoints control a single arm and cannot perform its two-arm tasks. On the real robot, we evaluate all methods ourselves and use the DROID-finetuned $\pi_{0.5}$ checkpoint.

\FloatBarrier
\section{Additional Experiment Results}
\label{sec:supp_pertask}

This section provides the complete benchmark results underlying the main comparison and organizes the remaining analyses around six research questions: the contributions of RoboAuto's components, the transferability of its hierarchical knowledge base, the effect of scaling knowledge-base entries, its performance with a stronger coding model, its sensitivity to search randomness, and its token efficiency.

\subsection{Generality and Transferability of Object-Centric Knowledge}
\label{sec:supp_demo_overlap}

The initial object-centric knowledge is constructed from LIBERO-90 demonstrations, and part of our evaluation uses the LIBERO-PRO and LIBERO-Plus benchmarks. The knowledge base is therefore designed to store physical understanding at a level of abstraction that transfers across tasks, objects, and scenes. Each entry describes \emph{how} a class of objects can be manipulated and \emph{what} must be established in the current scene, while the concrete program and its parameters are left to program search and scene grounding. We characterize this abstraction through four properties and verify that every entry satisfies them.

\paragraph{Verification procedure}
We read every entry sentence by sentence and checked it against the four properties below. To complement manual inspection, we applied pattern checks to every
sentence: (i)~\emph{numbers}, matching any digit sequence
(\texttt{[0-9]+(\textbackslash.[0-9]+)?}) and spelled-out numerals;
(ii)~\emph{units}, matching length, angle, mass, and force
(e.g., \texttt{mm}, \texttt{cm}, \texttt{m}, \texttt{deg}, \texttt{rad},
\texttt{\textdegree}, \texttt{kg}, \texttt{N}) following a number or
standing alone as a token; (iii)~\emph{task instructions}, matching each
instruction from all evaluation suites (LIBERO-PRO, LIBERO-PRO Long,
LIBERO-Plus, Robosuite, and RoboTwin~2.0) after lowercasing and punctuation
removal, both as a full phrase and as its verb--object sub-phrases
(e.g., ``put the bowl on the stove'' and ``bowl on the stove''); and
(iv)~\emph{API names}, matching the name of every perception and control
primitive exposed to the coder as well as any
function-call pattern (\texttt{[A-Za-z\_]+\textbackslash(}).
Every match was reviewed manually.

\subsubsection{Control Action Check}

Entries describe interaction strategies qualitatively, such as ``rim straddle away from the handle,'' and leave their implementation to program search. Manual inspection checked every sentence for commanded poses, joint angles, trajectories, gripper commands, and executable code, and the API-name pattern searched for every primitive name and any function-call syntax. No entry contains these elements.

\subsubsection{Physical Quantity Check}

Entries identify \emph{which} physical quantities matter for an object class through their \texttt{open\_quantities} field, such as the rim radius of a bowl, without recording values for particular objects or scene layouts. This keeps each entry applicable across object sizes and layouts, and the values are measured anew in every scene by Physical Scene Grounding (\cref{sec:ground}). The number and unit patterns confirmed that no entry stores a physical value.

\subsubsection{Task Instruction Check}

Knowledge is organized by object class rather than by task, so entries do not encode task instructions or ordered action sequences for a particular goal. We matched every sentence against all evaluation instructions and their verb--object sub-phrases, and manual inspection checked for task-level plans; no entry contains either.

\subsubsection{Object Instance Check}

Entries are indexed by general categories defined by affordances, spatial relations, and action effects, rather than by specific object instances such as ``cream cheese,'' ``wine bottle,'' or ``cookie box.'' Manual inspection of every \texttt{object\_class} and every sentence confirmed this organization. The complete list of categories follows.

\paragraph{Affordance}
\begin{itemize}[leftmargin=*, itemsep=0pt, topsep=2pt]
    \item Open-topped vessel wider than the finger opening (low, wide bowl without a handle)
    \item Open-topped vessel with a protruding side handle (mug)
    \item Upright short cylinder narrower than the finger opening (can)
    \item Upright bottle with a narrow cap or neck and a wider body
    \item Tall, narrow bottle with a long neck
    \item Small rigid box narrower than the finger opening
    \item Tall carton, upright or lying flat
    \item Thin slab standing on its edge (book)
    \item Wide, shallow pan with a long straight handle
    \item Upright pot with a protruding side handle
    \item Rotary knob with a raised ridge
    \item Sliding drawer (opening or closing)
    \item Hinged appliance door
    \item Flat, thin plate relocated without lifting
\end{itemize}

\paragraph{Spatial}
\begin{itemize}[leftmargin=*, itemsep=0pt, topsep=2pt]
    \item Placing a vessel or mug onto a flat plate
    \item Placing an item into a walled, open container
    \item Placing an item onto a raised flat surface
    \item Inserting a slab into a narrow, open-topped compartment
    \item Horizontal insertion into a side-opening shelf, onto a high top, or under a shelf
    \item Nesting one vessel into an identical vessel
    \item Relative placement beside or in front of a reference object
    \item Hanging pose of a held object and its propagation to placement
\end{itemize}

\paragraph{Effects}
\begin{itemize}[leftmargin=*, itemsep=0pt, topsep=2pt]
    \item Swept volume of the arm, the carried object, and moving articulations
    \item Clearing an articulation's swept volume in advance
    \item Articulation state and its visual indicators
    \item Release height and orientation changes upon release
    \item Unobservability caused by containment and occlusion
    \item Identical instances distinguishable only by spatial relations
    \item Absence of recovery evidence in the corpus
\end{itemize}

\subsubsection{Results by Similarity to the Knowledge Construction Data}

To examine how RoboAuto performs as evaluation tasks move further from the LIBERO-90 tasks used to construct the initial object-centric knowledge, we group all evaluation settings into three categories. Every evaluation task differs from the LIBERO-90 tasks, and the categories further differ in whether the scene layout and the set of manipulated objects also change: (1) same scene layout and same object set; (2) different scene layout, same object set; and (3) different scene layout and different object set. As shown in \Cref{tab:demo_overlap}, RoboAuto achieves the highest mean success rate in all three categories (74.5\%, 90.9\%, and 84.3\%), outperforming the strongest baseline by 45.5, 18.4, and 15.5 percentage points, respectively. Notably, RoboAuto maintains high performance in the third category, where both the scene layout and the object set are unseen during knowledge construction. This indicates that its gains do not stem from overlap with LIBERO-90, but from the transferable knowledge in the hierarchical knowledge base and the overall agent design.

\begin{table}[!tbp]
  \centering
  \begin{threeparttable}
  \caption{\textbf{Results by similarity to the LIBERO-90 demonstrations.} Mean success rate (\%) across evaluation settings, grouped by whether their scene layout and manipulated object set match those in the LIBERO-90 demonstrations used for knowledge construction. \#Set.\ denotes the number of evaluation settings in each row.}
  \label{tab:demo_overlap}
  \footnotesize
  \setlength{\tabcolsep}{2.5pt}
  \renewcommand{\arraystretch}{1.12}
  \begin{tabular}{@{}lrrrrrr@{}}
    \toprule
    Benchmark & \#Set. & RoboAuto & ASPIRE & \shortstack[r]{CaP-\\Agent0} & RHO & $\pi_{0.5}$ \\
    \midrule
    \multicolumn{7}{@{}l}{\textit{Same scene layout, same objects}} \\
    LIBERO-PRO Long & 10 & 73.4 & 30.8 & 8.6 & -- & 0.4 \\
    LIBERO-Plus & 2 & 80.0 & 20.0 & 1.0 & -- & 49.0 \\
    All & 12 & \textbf{74.5} & 29.0 & 7.3 & -- & 8.5 \\
    \midrule
    \multicolumn{7}{@{}l}{\textit{Different scene layout, same objects}} \\
    LIBERO-PRO Goal & 17 & 77.4 & 74.0 & 24.0 & 50.9 & 23.1 \\
    LIBERO-PRO Spatial & 20 & 94.8 & 55.3 & 12.9 & 57.3 & 10.6 \\
    LIBERO-PRO Object & 16 & 99.1 & 96.3 & 16.3 & 28.0 & 5.0 \\
    LIBERO-Plus & 4 & 95.5 & 57.0 & 49.5 & -- & 77.5 \\
    All & 57 & \textbf{90.9} & 72.5 & 19.7 & 46.4\tnote{a} & 17.4 \\
    \midrule
    \multicolumn{7}{@{}l}{\textit{Different scene layout, different objects}} \\
    LIBERO-PRO Goal & 3 & 52.7 & 1.3 & 5.3 & 65.3 & 2.0 \\
    LIBERO-PRO Object & 4 & 97.0 & 98.0 & 35.0 & 11.5 & 25.5 \\
    Robosuite & 7 & 90.6 & 81.0 & 68.3 & 70.0 & -- \\
    RoboTwin~2.0 & 4 & 91.5 & -- & -- & -- & -- \\
    Real robot\tnote{b} & 3 & 98.6 & 11.3 & -- & -- & 0.0 \\
    All\tnote{c} & 14 & \textbf{84.3} & 68.8 & 45.3 & 52.3 & 15.4\tnote{a} \\
    \bottomrule
  \end{tabular}
  \begin{tablenotes}[flushleft]\footnotesize
    \item[a] Mean over the settings in the category for which the method has results.
    \item[b] Stage-completion scores.
    \item[c] Mean over the 14 LIBERO-PRO and Robosuite settings. RoboTwin~2.0 is excluded because no baseline is evaluated on it, and the real robot is excluded because it reports stage-completion scores.
  \end{tablenotes}
  \end{threeparttable}
\end{table}

\subsubsection{Hierarchical Knowledge Base with ASPIRE}
\label{sec:kb_aspire}

To test whether the hierarchical knowledge base also improves an existing program-search method and whether it alone accounts for RoboAuto's gains, we add it to ASPIRE on ten short-horizon LIBERO-PRO tasks. As shown in \cref{tab:aspire_kb}, the knowledge base raises ASPIRE's success on three of the four Spatial tasks, from 8\% to 40\%, from 16\% to 40\%, and from 20\% to 38\%, while picking the bowl on the cookie box changes from 24\% to 18\%. On the three Object tasks, where ASPIRE already reaches 96\% to 100\%, success remains near ceiling, and on the three Goal tasks ASPIRE with the knowledge base scores 0\%. The knowledge base therefore transfers to a different search procedure but does not close the gap to RoboAuto, which reaches 84.8\% mean success on these tasks. On task-perturbed bowl placement on the stove, ASPIRE remains at 0\% with the hierarchical knowledge base while RoboAuto reaches 98\%, consistent with \cref{tab:ablation}, where the hierarchical knowledge base without scene grounding also yields 0\%. The knowledge base alone therefore does not reproduce RoboAuto's performance, which also depends on scene grounding to instantiate the entries in the current scene and on multi-stage planning with grounded verification and self-designed experiments to refine them.

\subsection{Complete Benchmark Results}
\label{sec:supp_complete_results}

\Cref{tab:supp_summary} summarizes performance across the five benchmark families. The following tables report every task and perturbation setting used to compute the aggregate results in the main paper.

\begin{table*}[tbp]
  \caption{\textbf{Mean success (\%) by benchmark and perturbation.} Each mean weights task--perturbation settings equally; setting counts are given under each column group. A dash denotes an unavailable result. The last row gives RoboAuto's difference from ASPIRE. Per-task results follow in \cref{tab:libero_pro_object,tab:libero_pro_goal,tab:libero_pro_spatial,tab:long,tab:plus,tab:robosuite}.}
  \label{tab:supp_summary}
  \centering
  \small
  \setlength{\tabcolsep}{2.3pt}
  \begin{tabular}{@{}l cccccc ccc ccc c@{}}
    \toprule
    & \multicolumn{6}{c}{LIBERO-PRO (60)} & \multicolumn{3}{c}{LIBERO-PRO Long (10)} & \multicolumn{3}{c}{LIBERO-Plus (6)} & Robosuite \\
    \cmidrule(lr){2-7}\cmidrule(lr){8-10}\cmidrule(lr){11-13}\cmidrule(lr){14-14}
    Method & Object & Goal & Spatial & Pos & Task & All & Pos & Task & All & Camera & Layout & All & (7) \\
    \midrule
    $\pi_{0.5}$~\cite{intelligence2025pi_} & 9.1 & 19.9 & 10.6 & 25.2 & 1.2 & 13.2 & 0.3 & 0.5 & 0.4 & 38.0 & \textbf{98.0} & 68.0 & -- \\
    CaP-Agent0~\cite{fu2026cap} & 20.0 & 21.2 & 12.9 & 19.7 & 16.3 & 18.0 & 9.3 & 7.5 & 8.6 & 24.0 & 42.7 & 33.3 & 68.3 \\
    RHO~\cite{elmaaroufi2026rho} & 24.7 & 53.1 & 57.3 & 38.6 & 51.5 & 45.0 & -- & -- & -- & -- & -- & -- & 70.0 \\
    ASPIRE~\cite{lu2026aspire} & 96.6 & 63.1 & 55.3 & 76.5 & 66.8 & 71.7 & 26.7 & 37.0 & 30.8 & 46.0 & 43.3 & 44.7 & 81.0 \\
    \midrule
    RoboAuto (ours) & \textbf{98.7} & \textbf{73.7} & \textbf{94.8} & \textbf{90.2} & \textbf{87.9} & \textbf{89.1} & \textbf{60.0} & \textbf{93.5} & \textbf{73.4} & \textbf{84.7} & 96.0 & \textbf{90.3} & \textbf{90.6} \\
    \quad\textit{vs.\ ASPIRE} & \textit{+2.1} & \textit{+10.6} & \textit{+39.5} & \textit{+13.7} & \textit{+21.1} & \textit{+17.4} & \textit{+33.3} & \textit{+56.5} & \textit{+42.6} & \textit{+38.7} & \textit{+52.7} & \textit{+45.7} & \textit{+9.6} \\
    \bottomrule
  \end{tabular}
\end{table*}

\subsubsection{Per-Task Results}

\Cref{tab:libero_pro} reports the LIBERO-PRO suite-level results. Detailed results for its Object, Goal, and Spatial suites are provided in \cref{tab:libero_pro_object,tab:libero_pro_goal,tab:libero_pro_spatial}. Results for LIBERO-PRO Long, LIBERO-Plus, and Robosuite appear in \cref{tab:long,tab:plus,tab:robosuite}, respectively.

\begin{table*}[!tbp]
  \centering
  \begin{threeparttable}
  \caption{\textbf{LIBERO-PRO suite means.} Equal-weight mean success (\%) across position and task perturbations.}
  \label{tab:libero_pro}
  \small
  \setlength{\tabcolsep}{4pt}
  \renewcommand{\arraystretch}{1.12}
  \begin{tabular}{@{}lrrrrr@{}}
    \toprule
    Suite & RoboAuto & ASPIRE & CaP-Agent0 & RHO & $\pi_{0.5}$ \\
    \midrule
    Object & 98.7 & 96.6 & 20.0 & 24.7 & 9.1 \\
    Goal & 73.7 & 63.1 & 21.2 & 53.1 & 19.9 \\
    Spatial & 94.8 & 55.3 & 12.9 & 57.3 & 10.6 \\
    \midrule
    All settings & 89.1 & 71.7 & 18.0 & 45.0 & 13.2 \\
    \bottomrule
  \end{tabular}
  \end{threeparttable}
\end{table*}

\begin{table*}[!tbp]
  \centering
  \begin{threeparttable}
  \caption{\textbf{LIBERO-PRO Object.} Reported success (\%) under position (Pos) and task (Task) perturbations; see \cref{tab:supp_protocol} for RoboAuto's seed protocol and search budget.}
  \label{tab:libero_pro_object}
  \small
  \setlength{\tabcolsep}{4pt}
  \renewcommand{\arraystretch}{1.12}
  \begin{tabular}{@{}rp{0.31\textwidth}rrrrrrrrrr@{}}
    \toprule
     & & \multicolumn{2}{c}{RoboAuto} & \multicolumn{2}{c}{ASPIRE} & \multicolumn{2}{c}{CaP-Agent0} & \multicolumn{2}{c}{RHO} & \multicolumn{2}{c}{$\pi_{0.5}$} \\
    \cmidrule(lr){3-4}\cmidrule(lr){5-6}\cmidrule(lr){7-8}\cmidrule(lr){9-10}\cmidrule(lr){11-12}
    ID & Task & Pos & Task & Pos & Task & Pos & Task & Pos & Task & Pos & Task \\
    \midrule
    1 & Place(alphabet\_soup,\allowbreak{}  basket) & 100 & 100 & 100 & 100 & 2 & 4 & 12 & 40 & 0 & 0 \\
    2 & Place(bbq\_sauce,\allowbreak{}  basket) & 92 & 98 & 100 & 100 & 12 & 42 & 32 & 88 & 100 & 2 \\
    3 & Place(butter,\allowbreak{}  basket) & 98 & 100 & 100 & 98 & 26 & 18 & 12 & 98 & 54 & 0 \\
    4 & Place(chocolate\_pudding,\allowbreak{}  basket) & 100 & 96 & 100 & 100 & 18 & 48 & 0 & 0 & 0 & 2 \\
    5 & Place(cream\_cheese,\allowbreak{}  basket) & 98 & 98 & 100 & 72 & 12 & 6 & 0 & 0 & 0 & 0 \\
    6 & Place(ketchup,\allowbreak{}  basket) & 98 & 100 & 92 & 96 & 32 & 12 & 0 & 4 & 20 & 2 \\
    7 & Place(milk,\allowbreak{}  basket) & 100 & 100 & 92 & 100 & 38 & 2 & 18 & 26 & 0 & 0 \\
    8 & Place(orange\_juice,\allowbreak{}  basket) & 98 & 98 & 100 & 100 & 30 & 2 & 34 & 40 & 0 & 2 \\
    9 & Place(salad\_dressing,\allowbreak{}  basket) & 100 & 100 & 96 & 92 & 32 & 0 & 14 & 76 & 0 & 0 \\
    10 & Place(tomato\_sauce,\allowbreak{}  basket) & 100 & 100 & 98 & 96 & 16 & 48 & 0 & 0 & 0 & 0 \\
    \midrule
    \multicolumn{2}{@{}l}{Mean} & 98.4 & 99.0 & 97.8 & 95.4 & 21.8 & 18.2 & 12.2 & 37.2 & 17.4 & 0.8 \\
    \bottomrule
  \end{tabular}
  \end{threeparttable}
\end{table*}

\begin{table*}[!tbp]
  \centering
  \begin{threeparttable}
  \caption{\textbf{LIBERO-PRO Goal.} Reported success (\%) under position (Pos) and task (Task) perturbations; see \cref{tab:supp_protocol} for RoboAuto's seed protocol and search budget.}
  \label{tab:libero_pro_goal}
  \small
  \setlength{\tabcolsep}{4pt}
  \renewcommand{\arraystretch}{1.12}
  \begin{tabular}{@{}rp{0.31\textwidth}rrrrrrrrrr@{}}
    \toprule
     & & \multicolumn{2}{c}{RoboAuto} & \multicolumn{2}{c}{ASPIRE} & \multicolumn{2}{c}{CaP-Agent0} & \multicolumn{2}{c}{RHO} & \multicolumn{2}{c}{$\pi_{0.5}$} \\
    \cmidrule(lr){3-4}\cmidrule(lr){5-6}\cmidrule(lr){7-8}\cmidrule(lr){9-10}\cmidrule(lr){11-12}
    ID & Task & Pos & Task & Pos & Task & Pos & Task & Pos & Task & Pos & Task \\
    \midrule
    1 & Open(cabinet,\allowbreak{}  drawer\_mid) & 100 & 4 & 72 & 0 & 0 & 0 & 0 & 0 & 0 & 4 \\
    2 & Put(bowl,\allowbreak{}  drawer\_top) & 0 & 26 & 86 & 46 & 4 & 0 & 30 & 32 & 94 & 2 \\
    3 & Push(plate,\allowbreak{}  stove\_front) & 0 & 6 & 0 & 76 & 0 & 10 & 0 & 0 & 0 & 0 \\
    4 & Put(bowl,\allowbreak{}  plate) & 100 & 100 & 92 & 88 & 36 & 38 & 98 & 98 & 0 & 2 \\
    5 & Put(bowl,\allowbreak{}  stove) & 98 & 98 & 96 & 0 & 22 & 12 & 100 & 98 & 0 & 4 \\
    6 & Put(bowl,\allowbreak{}  cabinet\_top) & 96 & 60 & 100 & 4 & 60 & 4 & 98 & 98 & 0 & 2 \\
    7 & Put(cream\_cheese,\allowbreak{}  bowl) & 92 & 100 & 74 & 74 & 4 & 34 & 10 & 10 & 98 & 2 \\
    8 & Put(wine\_bottle,\allowbreak{}  rack) & 100 & 100 & 98 & 86 & 2 & 12 & 58 & 62 & 88 & 2 \\
    9 & Put(wine\_bottle,\allowbreak{}  cabinet\_top) & 100 & 100 & 100 & 14 & 62 & 40 & 64 & 64 & 98 & 2 \\
    10 & TurnOn(stove) & 100 & 94 & 90 & 66 & 66 & 18 & 48 & 94 & 0 & 0 \\
    \midrule
    \multicolumn{2}{@{}l}{Mean} & 78.6 & 68.8 & 80.8 & 45.4 & 25.6 & 16.8 & 50.6 & 55.6 & 37.8 & 2.0 \\
    \bottomrule
  \end{tabular}
  \end{threeparttable}
\end{table*}

\begin{table*}[!tbp]
  \centering
  \begin{threeparttable}
  \caption{\textbf{LIBERO-PRO Spatial.} Reported success (\%) under position (Pos) and task (Task) perturbations; see \cref{tab:supp_protocol} for RoboAuto's seed protocol and search budget.}
  \label{tab:libero_pro_spatial}
  \small
  \setlength{\tabcolsep}{4pt}
  \renewcommand{\arraystretch}{1.12}
  \begin{tabular}{@{}rp{0.31\textwidth}rrrrrrrrrr@{}}
    \toprule
     & & \multicolumn{2}{c}{RoboAuto} & \multicolumn{2}{c}{ASPIRE} & \multicolumn{2}{c}{CaP-Agent0} & \multicolumn{2}{c}{RHO} & \multicolumn{2}{c}{$\pi_{0.5}$} \\
    \cmidrule(lr){3-4}\cmidrule(lr){5-6}\cmidrule(lr){7-8}\cmidrule(lr){9-10}\cmidrule(lr){11-12}
    ID & Task & Pos & Task & Pos & Task & Pos & Task & Pos & Task & Pos & Task \\
    \midrule
    1 & Pick(between(plate,\allowbreak{} ramekin),\allowbreak{}  plate) & 92 & 100 & 94 & 8 & 22 & 14 & 92 & 60 & 2 & 0 \\
    2 & Pick(table\_center,\allowbreak{}  plate) & 98 & 78 & 32 & 64 & 22 & 14 & 96 & 96 & 0 & 2 \\
    3 & Pick(drawer\_top(cabinet\_wood),\allowbreak{}  plate) & 78 & 92 & 10 & 96 & 2 & 10 & 6 & 4 & 0 & 0 \\
    4 & Pick(next\_to(cookie\_box),\allowbreak{}  plate) & 100 & 98 & 16 & 94 & 0 & 10 & 0 & 28 & 0 & 2 \\
    5 & Pick(next\_to(plate),\allowbreak{}  plate) & 78 & 98 & 100 & 70 & 10 & 20 & 56 & 64 & 0 & 0 \\
    6 & Pick(next\_to(ramekin),\allowbreak{}  plate) & 98 & 100 & 20 & 60 & 30 & 14 & 78 & 76 & 12 & 2 \\
    7 & Pick(on(cookie\_box),\allowbreak{}  plate) & 92 & 98 & 24 & 70 & 14 & 8 & 10 & 90 & 0 & 0 \\
    8 & Pick(on(ramekin),\allowbreak{}  plate) & 100 & 98 & 94 & 46 & 2 & 20 & 64 & 42 & 98 & 2 \\
    9 & Pick(on(stove),\allowbreak{}  plate) & 100 & 100 & 86 & 42 & 8 & 14 & 86 & 98 & 2 & 0 \\
    10 & Pick(on(cabinet\_wood),\allowbreak{}  plate) & 100 & 98 & 34 & 46 & 8 & 16 & 42 & 58 & 90 & 0 \\
    \midrule
    \multicolumn{2}{@{}l}{Mean} & 93.6 & 96.0 & 51.0 & 59.6 & 11.8 & 14.0 & 53.0 & 61.6 & 20.4 & 0.8 \\
    \bottomrule
  \end{tabular}
  \end{threeparttable}
\end{table*}

\begin{table*}[!tbp]
  \centering
  \begin{threeparttable}
  \caption{\textbf{LIBERO-PRO Long.} Reported success (\%); the seed protocol is in \cref{tab:supp_protocol}.}
  \label{tab:long}
  \small
  \setlength{\tabcolsep}{4pt}
  \renewcommand{\arraystretch}{1.12}
  \begin{tabular}{@{}rp{0.34\textwidth}lrrrr@{}}
    \toprule
    ID & Task & Perturbation & RoboAuto & ASPIRE & CaP-Agent0 & $\pi_{0.5}$ \\
    \midrule
    1 & Place(alphabet\_soup,\allowbreak{}  basket)\newline{}+ Place(cream\_cheese,\allowbreak{}  basket) & Pos & 100 & 46 & 40 & 0 \\
    2 & Put(white\_mug,\allowbreak{}  plate\_left)\newline{}+ Put(yellow\_white\_mug,\allowbreak{}  plate\_right) & Task & 100 & 60 & 12 & 2 \\
    3 & Put(white\_mug,\allowbreak{}  plate)\newline{}+ Put(chocolate\_pudding,\allowbreak{}  right\_of(plate)) & Task & 94 & 20 & 10 & 0 \\
    4 & TurnOn(stove)\newline{}+ Put(moka\_pot,\allowbreak{}  stove) & Task & 92 & 0 & 8 & 0 \\
    5 & Put(book,\allowbreak{}  caddy\_back) & Pos & 74 & 26 & 0 & 0 \\
    6 & Place(alphabet\_soup,\allowbreak{}  basket)\newline{}+ Place(tomato\_sauce,\allowbreak{}  basket) & Pos & 62 & 10 & 16 & 2 \\
    7 & Put(white\_mug,\allowbreak{}  plate\_left)\newline{}+ Put(yellow\_white\_mug,\allowbreak{}  plate\_right) & Pos & 58 & 22 & 0 & 0 \\
    8 & Put(white\_mug,\allowbreak{}  plate)\newline{}+ Put(chocolate\_pudding,\allowbreak{}  right\_of(plate)) & Pos & 66 & 54 & 0 & 0 \\
    9 & Put(moka\_pot\_1,\allowbreak{}  stove)\newline{}+ Put(moka\_pot\_2,\allowbreak{}  stove) & Pos & 0 & 2 & 0 & 0 \\
    10 & Put(book,\allowbreak{}  caddy\_back) & Task & 88 & 68 & 0 & 0 \\
    \midrule
    \multicolumn{3}{@{}l}{Mean} & 73.4 & 30.8 & 8.6 & 0.4 \\
    \bottomrule
  \end{tabular}
  \begin{tablenotes}[flushleft]\footnotesize
  \item[] Baseline configurations and result sources are described in Appendix~\ref{sec:supp_baselines}.
  \end{tablenotes}
  \end{threeparttable}
\end{table*}

\begin{table*}[!tbp]
  \centering
  \begin{threeparttable}
  \caption{\textbf{LIBERO-Plus.} Reported success (\%); the seed protocol is in \cref{tab:supp_protocol}.}
  \label{tab:plus}
  \small
  \setlength{\tabcolsep}{4pt}
  \renewcommand{\arraystretch}{1.12}
  \begin{tabular}{@{}rp{0.34\textwidth}lcrrrr@{}}
    \toprule
    ID & Task & Perturbation & Level & RoboAuto & ASPIRE & CaP-Agent0 & $\pi_{0.5}$ \\
    \midrule
    1 & Pick(next\_to(ramekin),\allowbreak{}  plate) & Camera & 5 & 88 & 10 & 46 & 10 \\
    2 & Pick(next\_to(ramekin),\allowbreak{}  plate) & Layout & 5 & 94 & 18 & 60 & 100 \\
    3 & Put(white\_mug,\allowbreak{}  plate)\newline{}+ Put(chocolate\_pudding,\allowbreak{}  right\_of(plate)) & Camera & 5 & 66 & 28 & 0 & 4 \\
    4 & Put(white\_mug,\allowbreak{}  plate)\newline{}+ Put(chocolate\_pudding,\allowbreak{}  right\_of(plate)) & Layout & 5 & 94 & 12 & 2 & 94 \\
    5 & TurnOn(stove) & Camera & 4 & 100 & 100 & 26 & 100 \\
    6 & TurnOn(stove) & Layout & 5 & 100 & 100 & 66 & 100 \\
    \midrule
    \multicolumn{4}{@{}l}{Mean} & 90.3 & 44.7 & 33.3 & 68.0 \\
    \bottomrule
  \end{tabular}
  \begin{tablenotes}[flushleft]\footnotesize
  \item[] Each setting uses the highest difficulty level available for its task and perturbation type: level~4 for the stove camera variation and level~5 for all other settings. Baseline configurations and result sources are described in Appendix~\ref{sec:supp_baselines}.
  \end{tablenotes}
  \end{threeparttable}
\end{table*}

\begin{table}[!htbp]
  \centering
  \begin{threeparttable}
  \caption{\textbf{Robosuite.} Reported success (\%) on seven tasks; see \cref{tab:supp_protocol} for seeds and budget.}
  \label{tab:robosuite}
  \small
  \setlength{\tabcolsep}{3pt}
  \renewcommand{\arraystretch}{1.12}
  \begin{tabular}{@{}r>{\raggedright\arraybackslash}p{0.30\columnwidth}rrrr@{}}
    \toprule
    ID & Task & RoboAuto & ASPIRE & \shortstack[r]{CaP-\\Agent0} & RHO \\
    \midrule
    1 & cube\_\allowbreak{}lift & 100 & 97 & 97 & 98 \\
    2 & cube\_\allowbreak{}stack & 98 & 99 & 98 & 98 \\
    3 & cube\_\allowbreak{}restack & 100 & 100 & 89 & 54 \\
    4 & spill\_\allowbreak{}wipe & 97 & 99 & 100 & 100 \\
    5 & two\_\allowbreak{}arm\_\allowbreak{}handover & 75 & 92 & 20 & 86 \\
    6 & two\_\allowbreak{}arm\_\allowbreak{}lift & 90 & 71 & 74 & 53 \\
    7 & nut\_\allowbreak{}assembly & 74 & 9 & 0 & 1 \\
    \midrule
    \multicolumn{2}{@{}l}{Mean} & 90.6 & 81.0 & 68.3 & 70.0 \\
    \bottomrule
  \end{tabular}
  \end{threeparttable}
\end{table}

\paragraph{Distribution of program success}
We further examine the distribution of final-program success across the 83 settings included in LIBERO-PRO, LIBERO-PRO Long, LIBERO-Plus, and Robosuite. This pool contains 60 LIBERO-PRO settings, 10 LIBERO-PRO Long settings, 6 LIBERO-Plus settings, and 7 Robosuite settings. RoboTwin~2.0 is excluded because it compares different RoboAuto knowledge-transfer conditions rather than RoboAuto with ASPIRE. Each final program is evaluated on held-out test scenes, 100 for Robosuite and 50 elsewhere.

Across these settings, RoboAuto achieves a mean success rate of 87.4\%, compared with 65.6\% for ASPIRE. The first quartile is 92.0\% for RoboAuto but only 30.0\% for ASPIRE, showing that the largest difference lies in the less reliably solved settings. RoboAuto reaches at least 90\% success on 64 of 83 settings (77\%), compared with 37 settings (45\%) for ASPIRE and 3 settings for CaP-Agent0.

The difference is especially pronounced on the 28 settings where ASPIRE achieves less than 50\% success. RoboAuto averages 76.6\% on this subset, compared with 19.5\% for ASPIRE, and reaches at least 80\% success on 17 of the 28 settings. On the 37 settings where ASPIRE already reaches at least 90\%, both methods average 97.3\%. The remaining 18 settings have ASPIRE success rates between 50\% and 90\%.

In direct per-setting comparisons, RoboAuto outperforms ASPIRE on 56 settings, ties on 9, and underperforms on 18. Using a descriptive tolerance of four percentage points, RoboAuto is higher on 46 settings, comparable on 31, and lower on 6. Against the strongest available baseline for each setting, whose average success is 76.6\%, RoboAuto matches or exceeds the baseline on 60 of the 83 settings.

\subsection{RQ1: How Does Each Component Contribute?}
\label{sec:supp_ablation}

\Cref{tab:ablation} evaluates the cumulative contributions of Search and Exploration, the Hierarchical Knowledge Base, and Physical Scene Grounding on five tasks. The base coding agent achieves only 5.2\% mean success. Adding Search and Exploration raises success to 19.2\%, including an improvement from 8\% to 82\% on the stove task. Sequential stage-wise search and independent verification localize incomplete interactions, while self-designed experiments collect targeted evidence about uncertain physical assumptions. Adding the Hierarchical Knowledge Base further increases mean success to 38.4\%. The mug-and-pudding task improves from 4\% to 78\%, illustrating how object-centric affordance knowledge can introduce an appropriate interaction strategy that repeated parameter revision does not discover. Finally, Physical Scene Grounding raises mean success to 91.2\%; drawer retrieval and bowl placement on the stove, which remain unsolved by the partial systems, reach 78\% and 98\%, respectively. These results show that the three components play complementary roles. Search and Exploration structures the improvement process, the knowledge base supplies reusable interaction strategies, and scene grounding instantiates those strategies for the current objects and geometry.

\begin{table*}[!tbp]
  \centering
  \begin{threeparttable}
  \caption{\textbf{Component ablation.} Reported success (\%). KB denotes the hierarchical knowledge base. Each task uses the seeds and budget of its suite in \cref{tab:supp_protocol}.}
  \label{tab:ablation}
  \small
  \setlength{\tabcolsep}{4pt}
  \renewcommand{\arraystretch}{1.12}
  \begin{tabular}{@{}l>{\raggedright\arraybackslash}p{0.34\textwidth}lrrrr@{}}
    \toprule
    Suite & Task & Perturbation & Base loop & + Search \& Exploration & + KB & RoboAuto \\
    \midrule
    Long & TurnOn(stove)\newline{}+ Put(moka\_pot,\allowbreak{}  stove) & Task & 8 & 82 & 84 & 92 \\
    Long & Put(white\_mug,\allowbreak{}  plate)\newline{}+ Put(chocolate\_pudding,\allowbreak{}  right\_of(plate)) & Task & 4 & 4 & 78 & 94 \\
    \begin{tabular}[t]{@{}l@{}}Spatial\\(\mbox{LIBERO-Plus})\end{tabular} & Pick(next\_to(ramekin),\allowbreak{}  plate) & Layout & 14 & 10 & 30 & 94 \\
    Spatial & Pick(drawer\_top(cabinet\_wood),\allowbreak{}  plate) & Pos & 0 & 0 & 0 & 78 \\
    Goal & Put(bowl,\allowbreak{}  stove) & Task & 0 & 0 & 0 & 98 \\
    \midrule
    \multicolumn{3}{@{}l}{Mean} & 5.2 & 19.2 & 38.4 & 91.2 \\
    \bottomrule
  \end{tabular}
  \begin{tablenotes}[flushleft]\footnotesize
  \item[] Columns add components cumulatively: Search and Exploration (verified stages and experiments), the Hierarchical Knowledge Base, then Physical Scene Grounding. The last column is the full system used in the main results.
  \end{tablenotes}
  \end{threeparttable}
\end{table*}

\begin{figure*}[t]
  \centering
  \includegraphics[width=\linewidth]{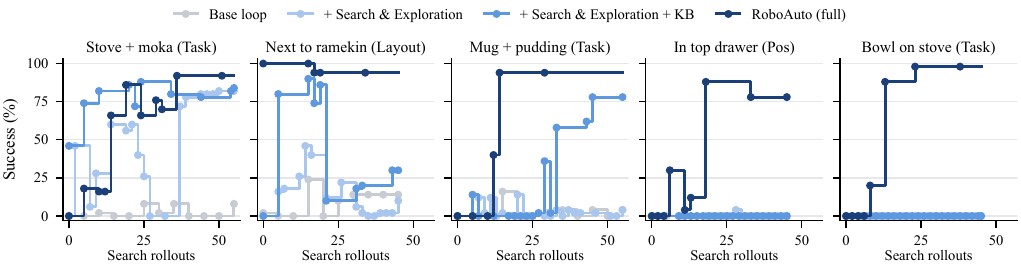}
  \caption{\textbf{Per-task ablation trajectories.} Test success versus the number of search rollouts as Search and Exploration, the \kb (KB), and scene grounding are added cumulatively to the base loop.}
  \label{fig:supp_ablation_curves}
\end{figure*}

\Cref{fig:supp_ablation_curves} shows the per-task search trajectories underlying \cref{fig:ablation}. \Cref{sec:supp_component_overview} illustrates how each added component changes the agent's behavior on the stove task.

\begin{table*}[!tbp]
  \centering
  \begin{threeparttable}
  \caption{\textbf{Hierarchical knowledge base with ASPIRE.} Reported success (\%). KB denotes the hierarchical knowledge base. Each task uses the seeds and budget of its suite in \cref{tab:supp_protocol}.}
  \label{tab:aspire_kb}
  \small
  \setlength{\tabcolsep}{4pt}
  \renewcommand{\arraystretch}{1.12}
  \begin{tabular}{@{}l>{\raggedright\arraybackslash}p{0.34\textwidth}lrrr@{}}
    \toprule
    Suite & Task & Perturbation & ASPIRE & ASPIRE + KB & RoboAuto \\
    \midrule
    Spatial & Pick(between(plate,\allowbreak{}  ramekin),\allowbreak{}  plate) & Task & 8 & 40 & 100 \\
    Spatial & Pick(next\_to(cookie\_box),\allowbreak{}  plate) & Pos & 16 & 40 & 100 \\
    Spatial & Pick(next\_to(ramekin),\allowbreak{}  plate) & Pos & 20 & 38 & 98 \\
    Spatial & Pick(on(cookie\_box),\allowbreak{}  plate) & Pos & 24 & 18 & 92 \\
    Goal & Put(bowl,\allowbreak{}  stove) & Task & 0 & 0 & 98 \\
    Goal & Open(cabinet,\allowbreak{}  drawer\_mid) & Task & 0 & 0 & 4 \\
    Goal & Put(bowl,\allowbreak{}  cabinet\_top) & Task & 4 & 0 & 60 \\
    Object & Place(chocolate\_pudding,\allowbreak{}  basket) & Task & 100 & 100 & 96 \\
    Object & Place(ketchup,\allowbreak{}  basket) & Task & 96 & 98 & 100 \\
    Object & Place(milk,\allowbreak{}  basket) & Task & 100 & 100 & 100 \\
    \midrule
    \multicolumn{3}{@{}l}{Mean} & 36.8 & 43.4 & 84.8 \\
    \bottomrule
  \end{tabular}
  \end{threeparttable}
\end{table*}

\subsection{RQ2: How Transferable Is the Hierarchical Knowledge Base?}
\label{sec:supp_transfer}

We evaluate RoboTwin~2.0 under the transfer conditions defined in Appendix~\ref{sec:supp_robotwin}.

Success rises from 24.5\% without retained knowledge to 91.5\% with the hierarchical knowledge base. The knowledge base transfers across the change in simulator and embodiment because its object-centric entries describe the physical requirements of an interaction, which RoboAuto re-grounds in the target scene and implements with the new robot's control primitives.

\begin{table}[!htbp]
  \centering
  \begin{threeparttable}
  \caption{\textbf{RoboTwin~2.0.} Reported success (\%) with an 11-rollout search budget and the seed protocol in \cref{tab:supp_protocol}.}
  \label{tab:supp_robotwin}
  \small
  \setlength{\tabcolsep}{3pt}
  \renewcommand{\arraystretch}{1.12}
  \begin{tabular}{@{}rp{0.30\columnwidth}rr@{}}
    \toprule
    ID & Task & \shortstack[r]{No retained\\knowledge} & \shortstack[r]{Hierarchical\\knowledge base} \\
    \midrule
    1 & Stack two bowls & 2 & 88 \\
    2 & Open microwave & 0 & 96 \\
    3 & Container on plate & 40 & 92 \\
    4 & Place empty cup & 56 & 90 \\
    \midrule
    \multicolumn{2}{@{}l}{Mean} & 24.5 & 91.5 \\
    \bottomrule
  \end{tabular}
  \begin{tablenotes}[flushleft]\footnotesize
  \item[] Both conditions use search and experiments.
  \end{tablenotes}
  \end{threeparttable}
\end{table}

\subsection{RQ3: Does Scaling Knowledge-Base Entries Improve Search?}
\label{sec:supp_kb_search}

We compare search with no hierarchical knowledge base, a randomly sampled half of the available entries, and the full knowledge base on five tasks. With no entries, scene grounding has nothing to measure, so the 0\% condition reuses the ``+ Search \& Exploration'' run of \cref{tab:ablation}; only the 50\% condition is new. \Cref{tab:kb_size} reports final-program success, while the search trajectories track test success as a function of the number of search rollouts. Results are averaged equally across tasks. When a rollout does not produce a new candidate program, the previous success is carried forward.

\begin{table*}[!tbp]
  \centering
  \begin{threeparttable}
  \caption{\textbf{Knowledge-base size.} Reported success (\%). KB denotes the hierarchical knowledge base. Each task uses the seeds and budget of its suite in \cref{tab:supp_protocol}.}
  \label{tab:kb_size}
  \small
  \setlength{\tabcolsep}{4pt}
  \renewcommand{\arraystretch}{1.12}
  \begin{tabular}{@{}l>{\raggedright\arraybackslash}p{0.34\textwidth}lrrr@{}}
    \toprule
    Suite & Task & Perturbation & 0\% KB & 50\% KB & 100\% KB \\
    \midrule
    Long & TurnOn(stove)\newline{}+ Put(moka\_pot,\allowbreak{}  stove) & Task & 82 & 86 & 92 \\
    Long & Put(white\_mug,\allowbreak{}  plate)\newline{}+ Put(chocolate\_pudding,\allowbreak{}  right\_of(plate)) & Task & 4 & 86 & 94 \\
    \begin{tabular}[t]{@{}l@{}}Spatial\\(\mbox{LIBERO-Plus})\end{tabular} & Pick(next\_to(ramekin),\allowbreak{}  plate) & Layout & 10 & 98 & 94 \\
    Spatial & Pick(drawer\_top(cabinet\_wood),\allowbreak{}  plate) & Pos & 0 & 56 & 78 \\
    Goal & Put(bowl,\allowbreak{}  stove) & Task & 0 & 0 & 98 \\
    \midrule
    \multicolumn{3}{@{}l}{Mean} & 19.2 & 65.2 & 91.2 \\
    \bottomrule
  \end{tabular}
  \begin{tablenotes}[flushleft]\footnotesize
  \item[] Each entry is the last program within the rollout budget, so that all three sizes follow one rule. With no entries, scene grounding has nothing to measure, so the 0\% column is the ``+ Search \& Exploration'' run of \cref{tab:ablation}.
  \end{tablenotes}
  \end{threeparttable}
\end{table*}

\begin{figure*}[t]
  \centering
  \includegraphics[width=\linewidth]{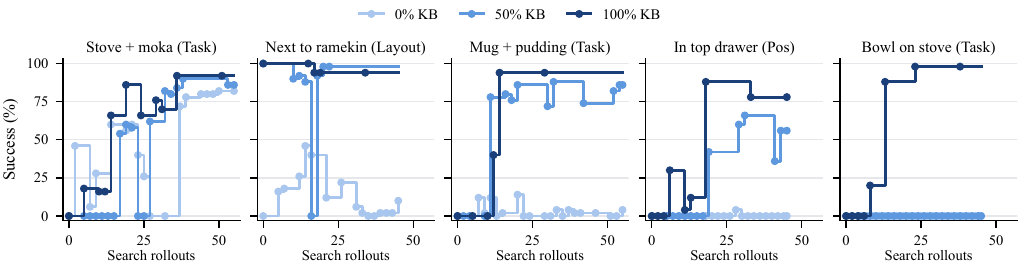}
  \caption{\textbf{Per-task knowledge-availability trajectories.} Test success versus the number of search rollouts with 0\%, randomly sampled 50\%, and 100\% of the \kb entries.}
  \label{fig:supp_kb_curves}
\end{figure*}

\Cref{fig:supp_kb_curves} shows the per-task search trajectories underlying \cref{fig:scaling}.

The three conditions achieve mean success rates of 21.6\%, 33.2\%, and 72.0\% after 15 rollouts. After 20 rollouts, they reach 22.0\%, 57.2\%, and 90.0\%, respectively, and end at 19.2\%, 65.2\%, and 91.2\%. The full and half knowledge bases begin at the same 20\% mean success, but their trajectories separate during revision, showing that additional entries help not only with the initial program but also with interpreting execution outcomes and selecting subsequent changes.

With the full knowledge base, every task first reaches at least 80\% success within 19 rollouts, at rollout indices 0, 13, 14, 18, and 19, with a median of 14; the drawer task's final program then scores 78\%. The half knowledge base reaches this threshold on three tasks, at rollout indices 0, 16, and 32, while search without object-centric knowledge reaches it on only one task, at rollout 44. Final-program success is 19.2\%, 65.2\%, and 91.2\% for the three conditions. Per task, the half knowledge base reaches 98\% final success on the ramekin task compared with 94\% for the full knowledge base, whereas bowl placement on the stove remains at 0\% with half of the entries and reaches 98\% with the full set. This experiment measures the effect of providing progressively broader knowledge subsets rather than attributing causal effects to individual entries.

\subsection{RQ4: Does RoboAuto Remain Effective with a Stronger Coder?}
\label{sec:supp_opus5}

We repeat five settings from the LIBERO-PRO Goal, Spatial, and Long suites with Opus~5 with a 1M-token context window as the coder for both the base loop and RoboAuto (\cref{tab:opus5}). RoboAuto reaches 77.6\% mean success against 29.6\% for the base loop, showing that its advantage holds with a stronger coder.

\paragraph{Stronger coding does not ensure physical understanding}
A stronger model such as Opus~5 has a stronger coding capability, and in particular a stronger ability to control the robot arm through the primitives. It writes its own kinematics when a primitive cannot reach a target, builds more of its own tooling, and, while evolving the program, attends to execution details that Opus~4.6 with a 1M-token context window misses, reading the rollout logs closely enough to catch primitive errors and offsets. These abilities let it reach 78\% on picking the bowl from the drawer in the base loop, since controlling the arm reliably and reading the logs carefully are enough to reach the bowl. However, on tasks that cannot be solved without grounded physical understanding, these abilities are not enough. On the long-horizon mug-and-pudding task it reaches 58\%, failing to pick the pudding reliably near the robot base; on plate pushing it reaches 12\%, never establishing how far each push moves the plate. On picking up the two moka pots and opening the cabinet drawer it fails entirely. It does not know how to grasp the moka pot reliably, its grasps repeatedly tip the pot over, and it spends its budget refining execution details and sweeping parameters from the feedback, such as the grasp height along the pot, the hand tilt angle and the closing-axis direction, without understanding what the problem is, how to interact with the object, or how to verify the result. The same unrecognized collision recurs when pulling out the cabinet drawer. Even with longer code and a precise inverse kinematics of its own, Opus~5 loses sight of the broader physical scene, and therefore spends its effort refining code that executes wrong physical assumptions well, with no chance to revise the understanding behind it.

\paragraph{RoboAuto grounds coding in physical understanding}
RoboAuto preserves the coder's execution capabilities while improving its understanding of physical constraints, allowing it to make revisions based on the geometry and mechanics of each scene and task. Before the first rollout, the coder identifies the intended interaction and measures the relevant physical quantities, such as the moka pot's handle apex or the clearance behind the drawer bar. These measurements let it revise the physical facts it has measured as well as the control parameters. Each failure is then attributed to its physical cause, because stage verification reads the object state after every step and active exploration of the environment tests what the rollouts leave unexplained, so the coder understands why a step failed instead of only that it failed. The moka placement fails and the stage check names a carry fault, so two turns of revision solve the task (0\% to 96\%); on plate pushing an experiment exposes the primitive's fingertip floor and a second one measures how far the plate travels per stroke below it (12\% to 74\%); on the cabinet drawer the coder establishes a pull along the measured slide axis with clearance from the cabinet, and its experiments trace the stalls to collisions with the cabinet, which the base loop had tuned around (0\% to 36\%); on the mug-and-pudding task the table floor near the robot base is measured and the finger placement verified before closing, which ends the blind retries (58\% to 84\%). RoboAuto therefore remains effective with a stronger coder.

\begin{table*}[!tbp]
  \centering
  \begin{threeparttable}
  \caption{\textbf{Stronger-model comparison.} Reported success (\%). Each task uses the seeds and budget of its suite in \cref{tab:supp_protocol}.}
  \label{tab:opus5}
  \small
  \setlength{\tabcolsep}{6pt}
  \renewcommand{\arraystretch}{1.12}
  \begin{tabular}{@{}l>{\raggedright\arraybackslash}p{0.50\textwidth}lrr@{}}
    \toprule
    Suite & Task & Perturbation & Base loop & RoboAuto \\
    \midrule
    Goal & Open(cabinet,\allowbreak{}  drawer\_mid) & Task & 0 & 36 \\
    Goal & Push(\allowbreak{}plate,\allowbreak{}  stove\_\allowbreak{}front) & Pos & 12 & 74 \\
    Spatial & Pick(\allowbreak{}drawer\_\allowbreak{}top(\allowbreak{}cabinet\_\allowbreak{}wood),\allowbreak{}  plate) & Pos & 78 & 98 \\
    Long & Put(\allowbreak{}white\_\allowbreak{}mug,\allowbreak{}  plate)\newline{}+ Put(\allowbreak{}chocolate\_\allowbreak{}pudding,\allowbreak{}  right\_\allowbreak{}of(\allowbreak{}plate)) & Pos & 58 & 84 \\
    Long & Put(\allowbreak{}moka\_\allowbreak{}pot\_\allowbreak{}1,\allowbreak{}  stove)\newline{}+ Put(\allowbreak{}moka\_\allowbreak{}pot\_\allowbreak{}2,\allowbreak{}  stove) & Pos & 0 & 96 \\
    \midrule
    \multicolumn{3}{@{}l}{Mean} & 29.6 & 77.6 \\
    \bottomrule
  \end{tabular}
  \begin{tablenotes}[flushleft]\footnotesize
  \item[] All runs use Opus~5 with a 1M-token context window.
  \end{tablenotes}
  \end{threeparttable}
\end{table*}

\subsection{RQ5: How Sensitive Is RoboAuto to Search Randomness?}
\label{sec:supp_variance}

To measure variation introduced by program search, we repeat the search on three randomly selected tasks. Run~1 is the main search run reported in the main results, and Runs~2 and~3 are two additional independent searches. Every resulting final program is evaluated on the same 50 test scenes, so differences across runs primarily reflect variation in the search process rather than variation in the test scenes. \Cref{tab:seeds} reports the success of each run together with the corresponding mean and sample standard deviation, providing a direct measure of the stability of the final programs discovered by RoboAuto.

\begin{table*}[!tbp]
  \centering
  \begin{threeparttable}
  \caption{\textbf{Repeated search runs.} Each final program is evaluated on test scenes (seeds 1--50). Mean and sample standard deviation are in percentage points.}
  \label{tab:seeds}
  \small
  \setlength{\tabcolsep}{6pt}
  \renewcommand{\arraystretch}{1.12}
  \begin{tabular}{@{}>{\raggedright\arraybackslash}p{0.43\textwidth}rrrrr@{}}
    \toprule
    Task / perturbation & Run 1 & Run 2 & Run 3 & Mean & Sample s.d. \\
    \midrule
    Mug + pudding,\allowbreak{}  task perturbed & 94 & 92 & 88 & 91.3 & 3.1 \\
    Pick(\allowbreak{}next\_\allowbreak{}to(\allowbreak{}ramekin),\allowbreak{}  plate),\allowbreak{}  camera perturbed & 88 & 92 & 86 & 88.7 & 3.1 \\
    Place(\allowbreak{}chocolate\_\allowbreak{}pudding,\allowbreak{}  basket),\allowbreak{}  task perturbed & 96 & 100 & 92 & 96.0 & 4.0 \\
    \bottomrule
  \end{tabular}
  \begin{tablenotes}[flushleft]\footnotesize
  \item[] Each search run uses the training and test intervals in \cref{tab:supp_protocol}.
  \end{tablenotes}
  \end{threeparttable}
\end{table*}

\subsection{RQ6: How Token-Efficient Is RoboAuto?}
\label{sec:supp_tokens}

We examine the relationship between program success and recorded inference usage. Because each method retains its own search procedure and stopping criteria, this analysis measures the token usage of the complete methods under their native configurations rather than under an artificially matched number of model calls. We report output and total token counts in millions and summarize per-task usage using the geometric mean, which reduces the influence of a small number of exceptionally long searches.

\begin{figure}[t]
  \centering
  \includegraphics[width=\linewidth]{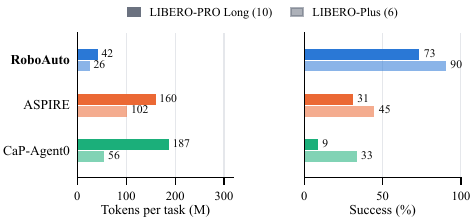}
  \caption{\textbf{Token usage and program success on matched tasks.}
  Geometric-mean total tokens per task and mean success on LIBERO-PRO Long and LIBERO-Plus settings with complete token records for RoboAuto, ASPIRE, and CaP-Agent0.}
  \label{fig:tokens}
\end{figure}

\paragraph{Matched comparison with autonomous Code-as-Policy baselines}
All sixteen LIBERO-PRO Long and LIBERO-Plus settings contain complete total-token records for RoboAuto, ASPIRE, and CaP-Agent0. On these settings, RoboAuto achieves 79.8\% mean success, compared with 36.0\% for ASPIRE and 17.9\% for CaP-Agent0. Its geometric-mean token usage is 35.1M per task, compared with 135.6M for ASPIRE and 118.8M for CaP-Agent0, so the two baselines use 3.9 and 3.4 times as many tokens as RoboAuto. The corresponding arithmetic means are 42.7M, 307.5M, and 168.6M tokens, giving ratios of 7.2 and 4.0.

As shown in \cref{fig:tokens}, RoboAuto more than doubles ASPIRE's mean success on both benchmarks while using approximately 26\% of its tokens on LIBERO-PRO Long and 25\% on LIBERO-Plus. It also achieves higher success with lower token usage than CaP-Agent0. Together with the search trajectories in \cref{sec:supp_kb_search}, these results show that RoboAuto's improvement does not require greater recorded inference usage. Its hierarchical knowledge base, stage-wise search, grounded verification, and targeted experiments collectively focus program revision on the interactions that remain unresolved.

\paragraph{RoboTwin~2.0 knowledge transfer}
The relationship between knowledge and token usage differs in the cross-embodiment transfer setting. The condition without retained knowledge uses an average of 9.3M total tokens per task and achieves 24.5\% success, whereas transferring the hierarchical knowledge base uses 23.6M tokens and achieves 91.5\% success. The knowledge base therefore substantially improves transfer, but the agent also performs more inference while grounding that knowledge and developing successful programs for the new simulator and embodiment.

\paragraph{Token-record coverage}
\Cref{tab:tokens_long,tab:tokens_plus,tab:tokens_robotwin} report per-task output and total token counts for LIBERO-PRO Long, LIBERO-Plus, and RoboTwin~2.0, the suites for which token records are available. A dash denotes an unreported value and is not treated as zero. Aggregate comparisons include only settings with complete output and total-token records for every method being compared, ensuring that each mean covers the same tasks. We retain the supplied total-token accounting without inferring an unavailable breakdown among input, cache, coder, auditor, or other model calls.

\begin{table}[H]
  \centering
  \begin{threeparttable}
  \caption{\textbf{RoboTwin~2.0 token usage.} Output and total counts are in millions. Task IDs correspond to \cref{tab:supp_robotwin}.}
  \label{tab:tokens_robotwin}
  \small
  \setlength{\tabcolsep}{3pt}
  \renewcommand{\arraystretch}{1.12}
  \begin{tabular}{@{}rp{0.30\columnwidth}rrrr@{}}
    \toprule
     &  & \multicolumn{2}{c}{\shortstack{No retained\\knowledge}} & \multicolumn{2}{c}{\shortstack{Hierarchical\\knowledge base}} \\
    \cmidrule(lr){3-4}\cmidrule(lr){5-6}
    ID & Task & \shortstack[r]{Output\\(M) $\downarrow$} & \shortstack[r]{Total\\(M) $\downarrow$} & \shortstack[r]{Output\\(M) $\downarrow$} & \shortstack[r]{Total\\(M) $\downarrow$} \\
    \midrule
    1 & Stack two bowls & 0.190 & 8.780 & 0.210 & 18.620 \\
    2 & Open microwave & 0.280 & 13.710 & 0.420 & 33.320 \\
    3 & Container on plate & 0.180 & 8.580 & 0.280 & 29.890 \\
    4 & Place empty cup & 0.080 & 6.170 & 0.120 & 12.640 \\
    \midrule
    \multicolumn{2}{@{}l}{Mean} & 0.183 & 9.310 & 0.258 & 23.618 \\
    \bottomrule
  \end{tabular}
  \begin{tablenotes}[flushleft]\footnotesize
  \item[] Mean rows use only tasks with complete output and total records for every compared method. Total counts reproduce the recorded accounting.
  \end{tablenotes}
  \end{threeparttable}
\end{table}

\begin{table*}[!tbp]
  \centering
  \begin{threeparttable}
  \caption{\textbf{LIBERO-PRO Long token usage.} Output and total counts are in millions. Task IDs correspond to \cref{tab:long}.}
  \label{tab:tokens_long}
  \small
  \setlength{\tabcolsep}{4pt}
  \renewcommand{\arraystretch}{1.12}
  \begin{tabular}{@{}rp{0.34\textwidth}lrrrrrr@{}}
    \toprule
     &  &  & \multicolumn{2}{c}{RoboAuto} & \multicolumn{2}{c}{ASPIRE} & \multicolumn{2}{c}{CaP-Agent0} \\
    \cmidrule(lr){4-5}\cmidrule(lr){6-7}\cmidrule(lr){8-9}
    ID & Task & Perturbation & \shortstack[r]{Output\\(M) $\downarrow$} & \shortstack[r]{Total\\(M) $\downarrow$} & \shortstack[r]{Output\\(M) $\downarrow$} & \shortstack[r]{Total\\(M) $\downarrow$} & \shortstack[r]{Output\\(M) $\downarrow$} & \shortstack[r]{Total\\(M) $\downarrow$} \\
    \midrule
    1 & Place(alphabet\_soup,\allowbreak{}  basket)\newline{}+ Place(cream\_cheese,\allowbreak{}  basket) & Pos & 0.200 & 15.700 & 0.240 & 39.080 & 25.500 & 176.200 \\
    2 & Put(white\_mug,\allowbreak{}  plate\_left)\newline{}+ Put(yellow\_white\_mug,\allowbreak{}  plate\_right) & Task & 1.160 & 89.270 & 0.763 & 2099.000 & 34.900 & 254.800 \\
    3 & Put(white\_mug,\allowbreak{}  plate)\newline{}+ Put(chocolate\_pudding,\allowbreak{}  right\_of(plate)) & Task & 0.610 & 26.030 & 0.420 & 162.430 & 51.600 & 276.100 \\
    4 & TurnOn(stove)\newline{}+ Put(moka\_pot,\allowbreak{}  stove) & Task & 0.390 & 28.490 & 0.690 & 300.850 & 35.800 & 443.700 \\
    5 & Put(book,\allowbreak{}  caddy\_back) & Pos & 0.520 & 34.880 & 0.300 & 171.080 & 40.600 & 465.300 \\
    6 & Place(alphabet\_soup,\allowbreak{}  basket)\newline{}+ Place(tomato\_sauce,\allowbreak{}  basket) & Pos & 0.730 & 64.960 & 0.150 & 95.320 & 18.900 & 369.900 \\
    7 & Put(white\_mug,\allowbreak{}  plate\_left)\newline{}+ Put(yellow\_white\_mug,\allowbreak{}  plate\_right) & Pos & 0.680 & 51.940 & 1.320 & 666.650 & 7.530 & 75.800 \\
    8 & Put(white\_mug,\allowbreak{}  plate)\newline{}+ Put(chocolate\_pudding,\allowbreak{}  right\_of(plate)) & Pos & 0.700 & 61.080 & 0.230 & 53.190 & 9.980 & 101.690 \\
    9 & Put(moka\_pot\_1,\allowbreak{}  stove)\newline{}+ Put(moka\_pot\_2,\allowbreak{}  stove) & Pos & 1.030 & 86.810 & 0.180 & 239.380 & 25.700 & 91.400 \\
    10 & Put(book,\allowbreak{}  caddy\_back) & Task & 0.360 & 28.200 & 0.120 & 20.340 & 7.500 & 79.410 \\
    \midrule
    \multicolumn{3}{@{}l}{Mean} & 0.638 & 48.736 & 0.441 & 384.732 & 25.801 & 233.430 \\
    \bottomrule
  \end{tabular}
  \begin{tablenotes}[flushleft]\footnotesize
  \item[] Mean rows use only tasks with complete output and total records for every compared method. Total counts reproduce the recorded accounting.
  \end{tablenotes}
  \end{threeparttable}
\end{table*}

\begin{table*}[!tbp]
  \centering
  \begin{threeparttable}
  \caption{\textbf{LIBERO-Plus token usage.} Output and total counts are in millions. Task IDs correspond to \cref{tab:plus}.}
  \label{tab:tokens_plus}
  \small
  \setlength{\tabcolsep}{4pt}
  \renewcommand{\arraystretch}{1.12}
  \begin{tabular}{@{}rp{0.34\textwidth}lrrrrrr@{}}
    \toprule
     &  &  & \multicolumn{2}{c}{RoboAuto} & \multicolumn{2}{c}{ASPIRE} & \multicolumn{2}{c}{CaP-Agent0} \\
    \cmidrule(lr){4-5}\cmidrule(lr){6-7}\cmidrule(lr){8-9}
    ID & Task & Perturbation & \shortstack[r]{Output\\(M) $\downarrow$} & \shortstack[r]{Total\\(M) $\downarrow$} & \shortstack[r]{Output\\(M) $\downarrow$} & \shortstack[r]{Total\\(M) $\downarrow$} & \shortstack[r]{Output\\(M) $\downarrow$} & \shortstack[r]{Total\\(M) $\downarrow$} \\
    \midrule
    1 & Pick(next\_to(ramekin),\allowbreak{}  plate) & Camera & 0.590 & 26.200 & 0.313 & 94.480 & 4.630 & 59.140 \\
    2 & Pick(next\_to(ramekin),\allowbreak{}  plate) & Layout & 0.420 & 39.920 & 0.509 & 624.260 & 2.260 & 33.280 \\
    3 & Put(white\_mug,\allowbreak{}  plate)\newline{}+ Put(chocolate\_pudding,\allowbreak{}  right\_of(plate)) & Camera & 0.570 & 38.500 & 0.137 & 106.450 & 9.910 & 95.580 \\
    4 & Put(white\_mug,\allowbreak{}  plate)\newline{}+ Put(chocolate\_pudding,\allowbreak{}  right\_of(plate)) & Layout & 1.530 & 70.240 & 0.156 & 34.180 & 9.710 & 91.310 \\
    5 & TurnOn(stove) & Camera & 0.160 & 8.640 & 0.152 & 29.210 & 3.310 & 44.900 \\
    6 & TurnOn(stove) & Layout & 0.280 & 11.800 & 0.323 & 183.990 & 3.000 & 38.390 \\
    \midrule
    \multicolumn{3}{@{}l}{Mean} & 0.592 & 32.550 & 0.265 & 178.762 & 5.470 & 60.433 \\
    \bottomrule
  \end{tabular}
  \begin{tablenotes}[flushleft]\footnotesize
  \item[] Mean rows use only tasks with complete output and total records for every compared method. Total counts reproduce the recorded accounting.
  \end{tablenotes}
  \end{threeparttable}
\end{table*}

\FloatBarrier
\section{Method Details}
\label{sec:supp_method}

This section provides the implementation details of RoboAuto's three components. We first describe how its hierarchical knowledge base is constructed and updated, then explain how physical knowledge is grounded in the current scene, and finally detail the stage-wise search, independent verification, and self-designed experimentation procedures.

\subsection{Hierarchical Knowledge Base}
\label{sec:supp_method_kb}

RoboAuto maintains two complementary forms of reusable experience: object-centric physical knowledge and executable skills. Object-centric entries describe transferable properties of physical interactions, while executable skills retain verified program implementations. Both layers continue to expand through online task solving.

\begin{figure*}[t]
  \centering
  \includegraphics[width=\linewidth]{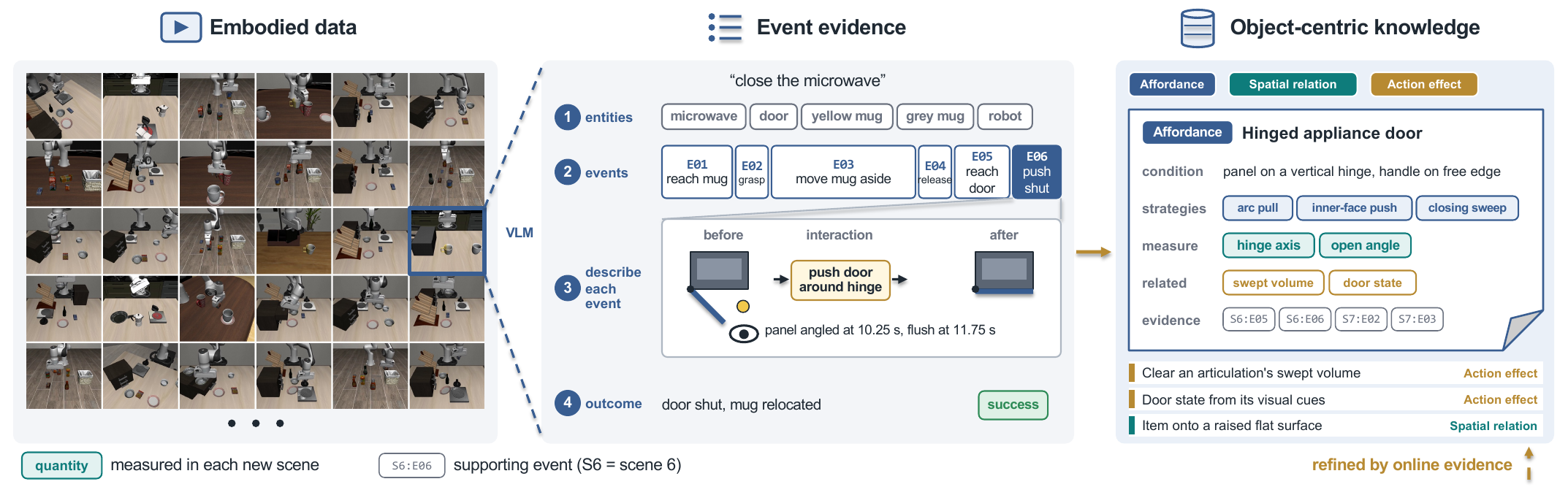}
  \caption{\textbf{Constructing object-centric knowledge from embodied data.}
  A vision-language model converts each demonstration video into an evidence record containing entities, temporally localized events, and an episode outcome. The example shows the LIBERO-90 task ``close the microwave,'' in which event E06 moves the door from open to flush. Knowledge distilled across such records is organized into affordances, spatial relations, and action effects. The expanded affordance entry cites supporting events from multiple episodes and declares the scene-dependent quantities, shown in teal, that must be measured again when the entry is applied in a new scene.}
  \label{fig:kb_construction}
\end{figure*}

\paragraph{Constructing object-centric knowledge from embodied data}
\Cref{fig:kb_construction} illustrates how the object-centric layer is constructed from embodied demonstration videos. A vision-language model converts each video into an evidence record through four steps. It first identifies the entities involved, segments the video into interaction events, describes each event using its preceding state, interaction, resulting state, and visual evidence, and finally evaluates the outcome of the episode. Object-centric knowledge is then distilled across these evidence records.

Each resulting entry is categorized as an affordance, spatial relation, or action effect. It specifies the relevant object class and applicability conditions, describes candidate interaction strategies, and declares the scene-dependent quantities required to instantiate them. The entry also links related knowledge and records the supporting demonstration events. During online task solving, new execution evidence can support, challenge, or refine these entries. \Cref{lst:kb_entries} shows one stored entry of each type. The knowledge field is abridged. Each of its sentences carries a tag stating whether it was demonstrated, inferred, or remains untested, and the open quantities name what must be measured in each scene before the entry is applied.

\onecolumn
\clearpage
\begin{lstlisting}[style=kb,basicstyle=\fontsize{8}{9}\selectfont\ttfamily,caption={One knowledge-base entry of each type.},label={lst:kb_entries}]
{"type": "affordance_constraints",
 "object_class": "Open-topped vessel with a protruding side handle (mug)",
 "strategies": ["RIM STRADDLE AWAY FROM HANDLE: RIM STRADDLE on the rim section diametrically or laterally away from the handle, so the handle never enters the finger gap.",
                "RIM STRADDLE BESIDE HANDLE: RIM STRADDLE on a rim section adjacent to the handle root, with the handle loop just outside one finger.",
                "HANDLE PINCH: close across the handle loop itself."],
 "knowledge": "Grasp mugs by the rim, not the handle. Prefer RIM STRADDLE AWAY FROM HANDLE; the loop visible immediately next to a finger in the wrist view, without collision. Select between them by the handle's bearing relative to the approach: if the rim section that faces the robot/carry direction contains the handle root, move the straddle point along the rim until the handle is outside the finger gap. HANDLE PINCH is untested for this class -- do not assume it; if tried, watch whether the mug swings about the handle when lifted. Carry the mug upright; it hangs level from the rim",
 "open_quantities": ["handle bearing around the rim", "rim radius"]}

{"type": "spatial_relationships",
 "object_class": "Slab into a narrow open-topped compartment",
 "strategies": ["IN-PLANE VERTICAL SLOT INSERTION: yaw the held slab so its plane matches the slot's long direction, position directly over the slot, and lower straight down."],
 "knowledge": "Use IN-PLANE VERTICAL SLOT INSERTION with the TOP-EDGE COVER PINCH kept throughout. Align yaw during the carry, before reaching the caddy, not at the slot. Map the compartment name to geometry from the caddy layout, not from image position alone: 'front'/'back' referred to the middle column's front and back slots; 'right' was the tall back-right slot; 'left' the long left slot. The slab ends standing with its upper part above the walls, hand still on the top edge; release is not shown.",
 "open_quantities": ["slot width and long axis", "insertion depth"]}

{"type": "effects_dependencies",
 "object_class": "Articulated state and its visual indicators",
 "strategies": [],
 "knowledge": "Read states from these cues: drawer OPEN = its slatted/dark interior box protrudes from the cabinet outline; CLOSED = front flush, interior invisible. Door OPEN = windowed panel standing out from the body; CLOSED = panel flush against the body. Stove ON = burner coil red; OFF = grey/dark. States persist without contact: a partly pulled drawer stays where left between strokes, a partly opened door stays at its angle when the hand repositions, a burner stays red after the hand leaves the knob. So a later stage may assume the articulation remains where last observed if nothing touched it. Closing a drawer with an item inside did not visibly disturb the item, but it becomes unobservable.",
 "open_quantities": []}
\end{lstlisting}
\twocolumn

\paragraph{Executable skills and online updates}
The executable layer stores program components that have been successfully verified across the current search pool. Unlike object-centric entries, these skills directly implement behaviors using the perception and control primitives of a particular robot interface. After a task is solved, verified program components are consolidated into the skill library, while the physical evidence collected during search updates the corresponding object-centric entries. This separation allows RoboAuto to retain both transferable physical knowledge and reusable executable implementations.

\subsection{Physical Scene Grounding}
\label{sec:supp_method_ground}

Physical Scene Grounding determines how retrieved object-centric knowledge applies to the current task and scene. Given the task instruction and initial observations, RoboAuto identifies the task-relevant objects, associates them with retrieved knowledge entries, and checks the entries' applicability conditions. It then establishes the action-critical quantities declared by those entries, such as object centers, dimensions, orientations, grasp locations, handle directions, and spatial relations.

RoboAuto combines visual reasoning and segmentation tools, including Molmo2 and SAM~3, with metric observations from scene and wrist depth cameras. The resulting task-specific grounding record stores the quantities that have been established, their supporting observations, and any conditions that remain unresolved. The coder uses this record to select feasible strategies and instantiate their parameters. Grounding remains revisable during search: an execution outcome may challenge an earlier estimate, reveal an overlooked condition, or require an additional measurement or targeted experiment.

\subsection{Search and Exploration}
\label{sec:supp_method_search}

\begin{figure*}[t]
  \centering
  \includegraphics[width=\linewidth]{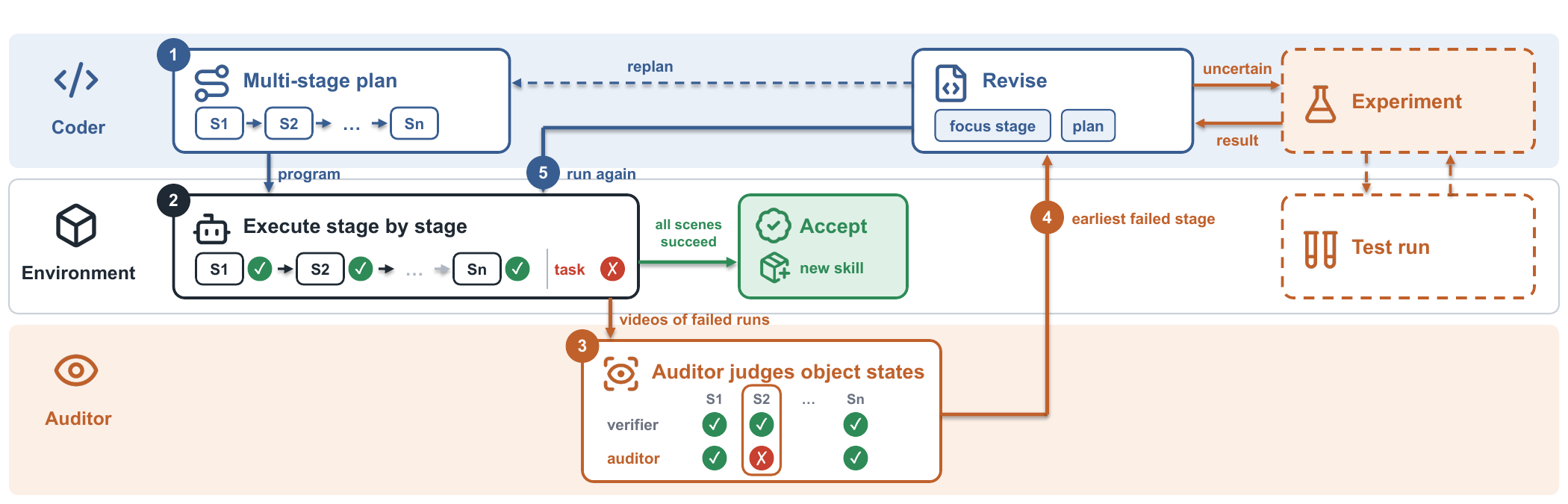}
  \caption{\textbf{Search and Exploration.}
  The coder decomposes the task into stages with intended object states and executable verifiers (1), and the environment executes these stages sequentially (2). When an execution fails, an independent auditor evaluates the stage outcomes from synchronized videos and may reject a state that the program verifier accepted (3). The earliest incomplete stage is returned to the coder (4), which revises the stage or replans the task before executing the next candidate program (5). If the cause remains uncertain, the coder branches into a separately implemented experiment that collects targeted physical evidence before revision continues. Programs verified across the search pool contribute reusable executable skills.}
  \label{fig:search_workflow}
\end{figure*}

Search and Exploration coordinates three roles: the coder generates and revises programs, the environment executes them, and an independent auditor reviews failed executions. As summarized in \cref{fig:search_workflow}, RoboAuto searches sequentially over task stages, verifies their physical outcomes, and designs targeted experiments when execution evidence does not explain a failure.

\paragraph{Multi-stage planning and sequential search}
The coder decomposes the task into stages, each defined by an intended change in object state and an executable verifier for that change. The stages are ordered according to their physical dependencies and executed sequentially. A failed verifier stops the remaining task stages, preventing later actions from obscuring or compounding an earlier failure.

Program revision focuses on the earliest stage whose intended state has not been established. Stages that already succeed can therefore be retained while search concentrates on the unresolved interaction. The coder may revise the stage implementation, its verifier, or the broader plan. If execution evidence shows that the current decomposition or dependency structure is invalid, RoboAuto can redesign the stages and replan the task rather than continuing to revise an unrealizable stage.

\paragraph{Independent grounded verification}
Program-generated verifiers may themselves be incorrect, so RoboAuto uses a separate language-model session to audit failed executions. The auditor receives the task instruction, stage plan, relevant physical knowledge, execution logs, and synchronized scene and wrist videos. It independently determines whether each stage produced its intended object state and describes the observed interaction.

When the auditor returns a judgment for a stage, that judgment replaces the corresponding program-verifier verdict. If the auditor provides no judgment for a stage, RoboAuto retains the program verifier's original result. The earliest stage judged incomplete becomes the focus of the next revision. Previously verified stages may also be reopened when later evidence reveals a regression. The environment's task predicate remains the final criterion for task-level success: successful scenes bypass video review, while a verifier failure reported on a successful scene is recorded as a verification error.

\paragraph{Self-designed experiments}
Recognizing an incomplete stage does not necessarily reveal which physical assumption caused the failure. When the available evidence is insufficient, RoboAuto formulates competing physical hypotheses and predicts the observations that would support them. It then writes a separate experiment program and corresponding control code to test these predictions. These files are distinct from the candidate robot program because their purpose is to acquire diagnostic evidence rather than complete the task.

Each experiment specifies a hypothesis, a predicted observation, an optional starting-state prefix, and controlled test actions. For example, RoboAuto may generate an experiment that evaluates several candidate grasp poses under the same scene conditions and compares whether each pose establishes stable contact. Prefix stages prepare the required starting state before the test actions. A failed prefix verdict remains visible to the agent but does not prevent the diagnostic actions from running, whereas an execution exception terminates the experiment. The returned logs and camera observations are interpreted relative to the observed starting state and recorded in a persistent diagnosis notebook. The evidence then informs program revision and may update the corresponding object-centric entry.

\subsubsection{Execution and Revision Protocol}
\label{sec:supp_prompts}

Each new or revised candidate is first executed on two scenes, prioritizing scenes in which the current focus stage previously failed. Once that stage succeeds on the initial batch, the same program is evaluated on the remainder of the current search pool. Solving the complete pool expands it from five to ten and then fifteen scenes. If the rollout budget is exhausted while failures remain, the coder produces a final evidence-guided revision.

The grounded knowledge record and diagnosis notebook persist across search iterations. Once the task is solved across the search pool, verified programs are retained as executable skills and relevant physical evidence is consolidated into the object-centric knowledge base.

\FloatBarrier
\section{Behavioral Analysis of RoboAuto}
\label{sec:supp_behavior}
\label{sec:supp_failures}
\label{sec:supp_cases}
\label{sec:supp_ablation_cases}

This section analyzes how RoboAuto behaves during program search and how its components change that behavior. We use two complementary forms of evidence. First, baseline case studies examine runs in which program revisions continue without resolving the physical conditions required by an interaction, and compare them with RoboAuto runs on the same task. Second, controlled ablations add Search and Exploration, the Hierarchical Knowledge Base, and Physical Scene Grounding cumulatively while holding the task, rollout budget, and scene seeds fixed.

The cases cover three recurring difficulties. They concern selecting a physically appropriate interaction strategy, grounding the quantities required by that strategy, and verifying that an executed action produced its intended physical effect. Frames are taken from the cited trials. Code and knowledge excerpts are condensed from the generated programs and stored entries while retaining their original identifiers and values.

\subsection{Behavior Across Component Ablations}
\label{sec:supp_component_overview}

The component ablation in \cref{tab:ablation} compares four cumulative variants on five tasks. We illustrate the progression on the stove task. The base loop revises complete programs from task-level rollout outcomes. The second variant adds Search and Exploration, including stage decomposition, grounded verification, and self-designed experiments. The third adds the Hierarchical Knowledge Base, and the final variant additionally grounds its object-centric entries in the current scene.

\begin{figure*}[p]
  \centering
  \includegraphics[width=\linewidth]{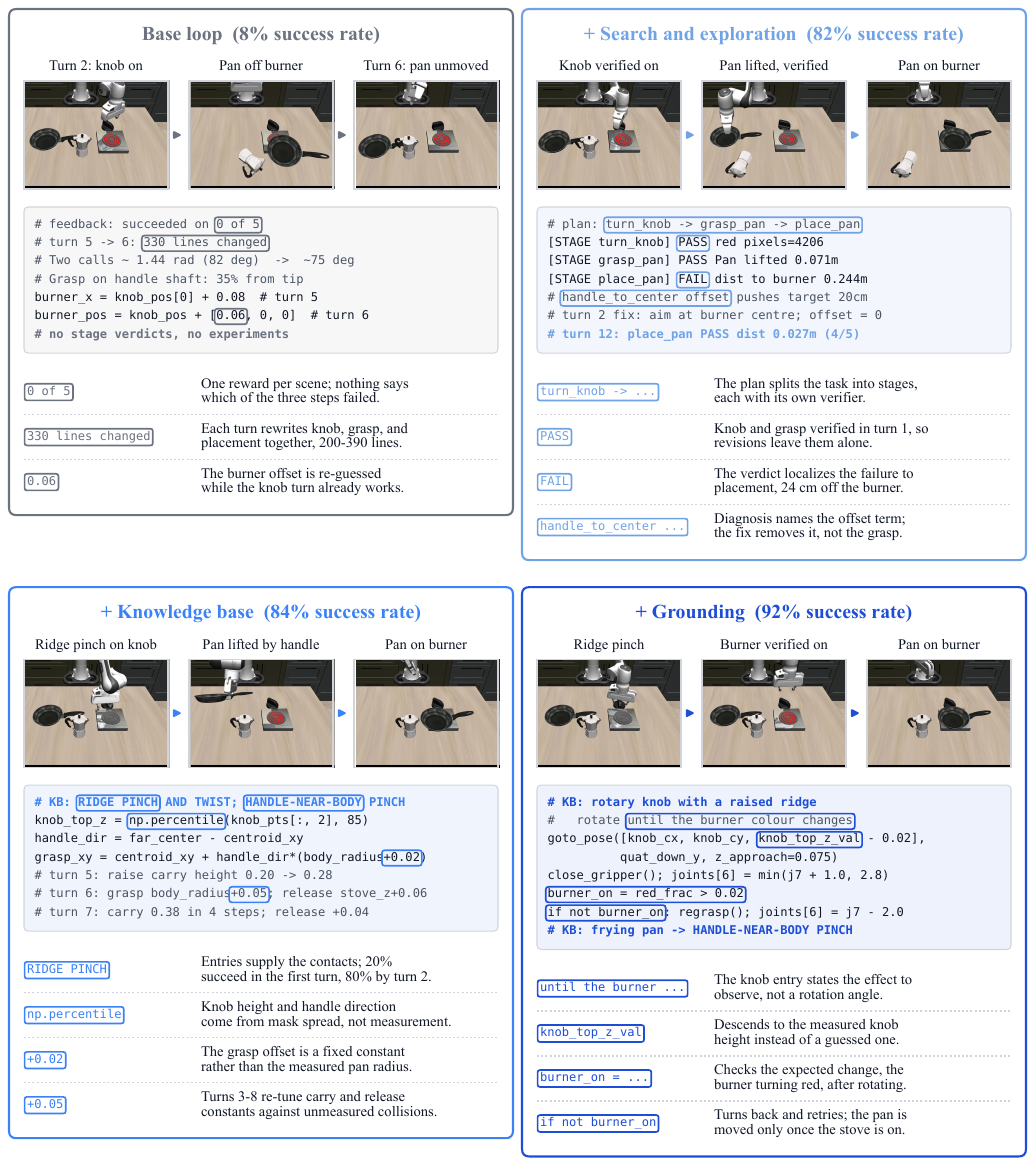}
  \caption{\textbf{The four component variants on turning on the stove and placing the pan (LIBERO-PRO Long, task perturbation of turning on the stove and placing the moka pot).}
  Each panel adds one component to the preceding variant. The base loop rewrites the complete program from a per-scene reward. Search and Exploration plans stages whose verdicts localize the failure. The Hierarchical Knowledge Base supplies knob and handle contacts in the first turn but estimates their quantities from segmentation-mask spread and fixed offsets. Physical Scene Grounding measures the knob height and pan geometry and verifies burner activation before the pan is moved.}
  \label{fig:components_stove}
\end{figure*}

\Cref{fig:components_stove} compares the four variants on a single task. Across nine turns, the base loop succeeds on only one training scene and receives no information about which interaction caused the remaining failures. Search and Exploration verifies the knob stage in its first turn and the pan grasp in its second, allowing subsequent revisions to concentrate on placement. Adding the knowledge base supplies workable knob and handle contacts in the first turn, but later revisions must tune unmeasured carry and release constants. The full system measures the pan, knob, and stove before generating its first program. By turn~9, after 36 candidate-program rollouts, it solves 14 of the 15 training scenes. Each component changes the point at which search stops rediscovering the interaction and begins refining its execution.

The remaining four ablation tasks in \cref{tab:ablation} follow the same progression. Without stages, the base loop rewrites the complete program from a per-scene reward and does not establish which interaction failed. Search and Exploration localizes the failed stage, but when that stage admits several explanations it may investigate an incorrect hypothesis first. The Hierarchical Knowledge Base removes much of the search over interaction families, and every task with a matching entry passes its grasp stage within the first one or two turns. What remains unresolved are the scene-dependent quantities required by the entry, which the knowledge-base-only variant approximates from segmentation-mask spread and fixed offsets. These approximations can place the gripper outside a rim or overshoot a placement. Physical Scene Grounding obtains these quantities from geometric estimators cross-checked across cameras and evaluates whether a stored strategy applies. When no stored strategy is feasible, RoboAuto uses a self-designed experiment to establish a new interaction.

\subsection{Hierarchical Knowledge Base:
Discovering Suitable Interaction Strategies}
\label{sec:supp_case_strategy}

The book case illustrates how object-centric knowledge supplies a suitable contact strategy when repeated parameter revisions preserve an unsuccessful grasp.

\subsubsection{An Unsuitable Book Grasp Persists Across Revisions}

\begin{figure*}[tp]
  \centering
  \includegraphics[width=0.9\linewidth]{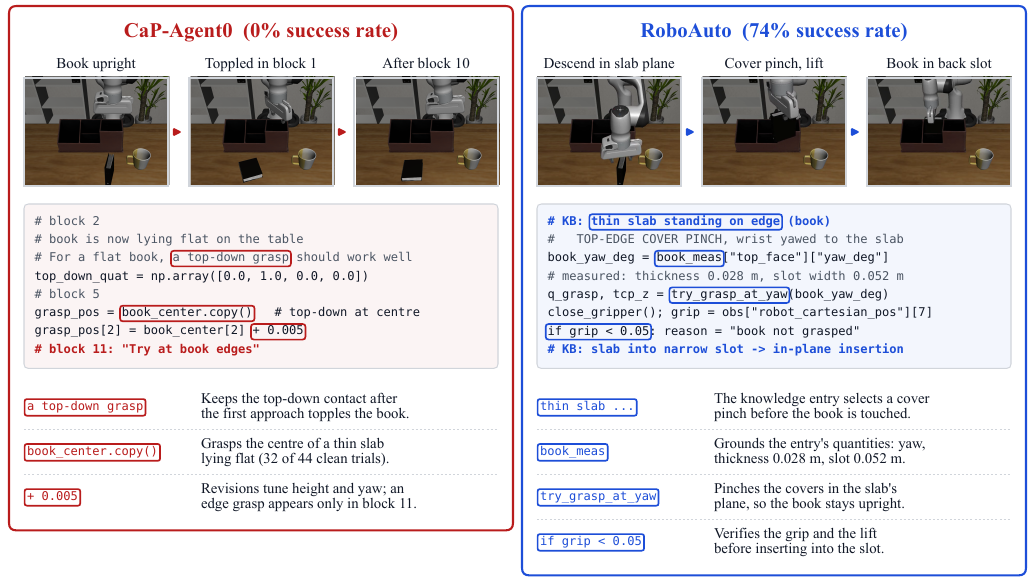}
  \caption{\textbf{Book into the caddy (LIBERO-PRO Long, position perturbation).}
  CaP-Agent0 continues using top-down grasps after its first approach topples the book. RoboAuto selects a cover pinch from the thin-slab entry and grounds the quantities required to execute it.}
  \label{fig:case_book}
\end{figure*}

\begin{figure*}[tp]
  \centering
  \includegraphics[width=0.9\linewidth]{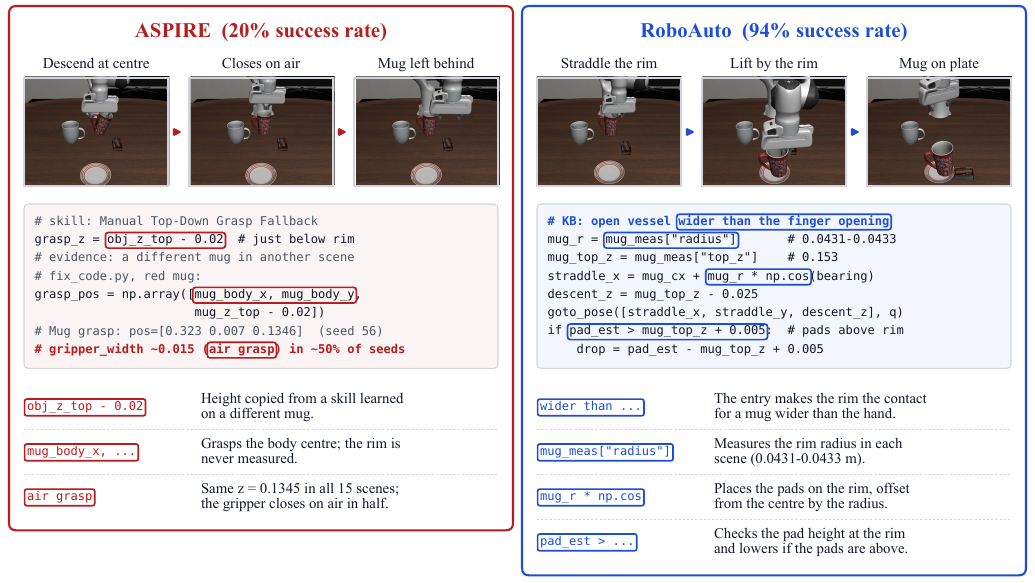}
  \caption{\textbf{Placing the red mug on the plate and the pudding to its right (LIBERO-PRO Long, task perturbation of placing the white mug on the plate and the pudding to its right).}
  ASPIRE applies a grasp-height rule stored in a skill learned on a different mug, causing the gripper to close on air in approximately half of the training scenes. RoboAuto measures the rim of the current mug and places the pads on it.}
  \label{fig:case_mugz}
\end{figure*}

\begin{figure*}[tp]
  \centering
  \includegraphics[width=0.9\linewidth]{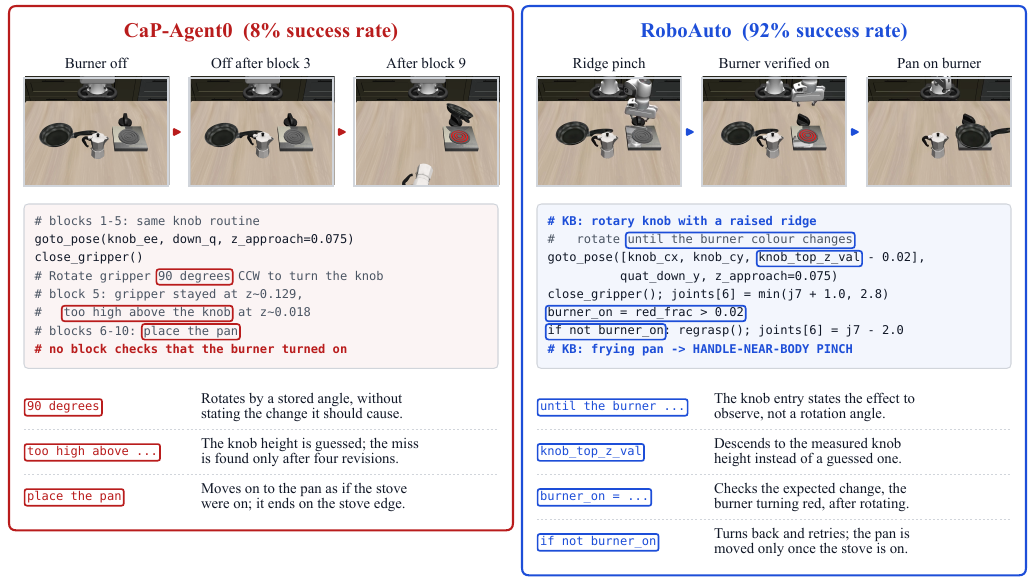}
  \caption{\textbf{Turning on the stove and placing the pan (LIBERO-PRO Long, task perturbation of turning on the stove and placing the moka pot).}
  CaP-Agent0 rotates the knob by a stored angle and proceeds without checking the burner. RoboAuto's knob entry specifies the expected observable effect, which the program verifies before beginning pan placement.}
  \label{fig:case_stove}
\end{figure*}

The book begins standing on its edge, as shown in \cref{fig:case_book}. In CaP-Agent0's trial~02, the first top-down approach knocks it flat. The following code block observes that the book is now lying on the table but concludes that another top-down grasp should work. Blocks~2--10 continue adjusting grasp height and orientation around the book center, and only block~11 of 12 attempts an edge grasp.

Across the 44 trials without service errors, every generated program uses top-down grasps, 32 grasp at the book center, 13 explicitly note that the book has fallen, and none completes the task. RoboAuto instead retrieves an entry for manipulating a thin slab standing on its edge. The entry prescribes a top-edge cover pinch supported by 14 book-grasp demonstrations and declares the book yaw and thickness as quantities to measure. RoboAuto obtains a thickness of 2.8\,cm, verifies consistency across the two cameras, and checks that the 5.2\,cm caddy slot can receive it. The program pinches the covers before the book is disturbed, verifies the gripper width and the displacement of the book, and inserts it along the slab plane, reaching 74\% success on held-out test scenes.

The difference lies in the interaction strategy rather than its parameters. Once the top-down grasp has toppled the book, further adjustment of height or orientation within that contact family does not recover the task.

\subsection{Physical Scene Grounding:
Instantiating Knowledge in the Current Scene}
\label{sec:supp_case_ground}

The mug case examines the effect of replacing an inherited grasp-height offset with measurements of the current object's rim.

\subsubsection{A Grasp-Height Rule Is Copied from Another Object}

ASPIRE's skill library contains a manually implemented top-down grasp with a fixed offset below the object top:

\[
\texttt{grasp\_z = obj\_z\_top - 0.02}.
\]

The rule was distilled from a run involving a different mug in another scene. ASPIRE reuses it for the red mug and grasps at the body center using the stored offset. Because the same mug model appears across scenes, the resulting grasp height is fixed at $z=0.1345$\,m in all 15 training scenes. The program's own findings report that this grasp produces a gripper width of approximately 0.015\,m, indicating an air grasp, in approximately half of the seeds, while the same code succeeds in others.

The stored rule contains no applicability condition that explains the variation. It does not state the rim radius, finger opening, or geometric relationship for which the offset was valid. ASPIRE's repair loop ends at 0/50 success; its reported 20\% on this setting is obtained through its subsequent evolutionary search.

RoboAuto's entry for an open vessel wider than the finger opening declares the rim radius and top height as quantities to measure. Across test trials, it obtains a radius between 0.0431 and 0.0433\,m and a top height of approximately 0.153\,m. The program offsets the pads from the mug center by the measured radius and checks the predicted pad height against the rim before closing, lowering the hand if the pads remain above it. The quantities vary little across these scenes, but they are explicitly tied to and verified against the object being grasped rather than inherited from a procedure that succeeded elsewhere.

\subsection{Search and Exploration:
Localizing and Investigating Failures}
\label{sec:supp_behavior_search}

The stove case illustrates why program search must verify the burner's state before proceeding to pan placement.

\subsubsection{An Executed Action Is Mistaken for an Achieved Physical State}
\label{sec:supp_case_state}

The instruction is to turn on the stove and place the pan on it. In CaP-Agent0's trial~18, blocks~1--5 repeat the same routine: descend toward the detected knob point, close the gripper, and rotate the wrist by 90 degrees. Block~5 records that the gripper remains at $z\approx0.129$\,m, far above the estimated knob height of $z\approx0.018$\,m. Nevertheless, no block checks whether the burner changed state.

The burner remains off through the first four revisions. Later blocks proceed to pan placement, and the pan ends propped against the stove edge. Across the failed trials, most programs infer the knob geometry from its expected position; only 7 of 93 reason explicitly about the knob ridge or its orientation.

RoboAuto's knob entry prescribes a ridge pinch and twist and specifies the effect to observe: rotate until the burner appearance changes, rather than by a stored angle. The program descends to the grounded knob height, performs the rotation, and measures the burner's red fraction. If the burner remains off, it re-grasps and rotates in the opposite direction. It grasps the pan near the handle root only after burner activation has been verified, reaching 92\% success on test scenes. Executing the rotation is not equivalent to achieving the intended physical state. Verifying the state localizes the failure to the knob stage rather than allowing it to appear only as a final reward of zero.

\subsection{Summary of Observed Behaviors}
\label{sec:supp_behavior_summary}

The cases illustrate distinct sources of progress during program search: object-centric knowledge supplies a cover-pinch strategy for the standing book, scene grounding replaces the mug's inherited grasp-height offset with measured rim geometry, and stage-wise verification checks burner activation before pan placement. The stove ablation further shows how these components change the revisions made during search, from discovering workable contacts to measuring and refining their execution.

\onecolumn
\clearpage
\section{Prompts}
\label{sec:supp_prompt_listings}

This section reproduces the prompts that drive program search. The coder prompt is sent once at the start of a task and remains in the session. The auditor prompt starts a fresh session for each audit. Per-turn feedback is assembled from the execution digest, the auditor grid, and a closing instruction. Workspace paths, file-handling instructions, and the primitive API documentation inherited from CaP-X~\cite{fu2026cap} are abridged.

\subsection{Coder Session Prompt}
\label{sec:supp_prompt_coder}

\Cref{lst:prompt_coder} shows the prompt that opens the coder session. It defines the scene pool and batching rules, the stage-file format with per-stage verifiers, the knowledge-base lookup and measurement requirements, the experiment protocol, and the lab-notebook discipline. The placeholder GOAL is replaced with the task instruction.

\begin{lstlisting}[style=prompt,caption={Coder session prompt.},label={lst:prompt_coder}]
===== THE TASK =====
You are controlling a robot with API described below.
Goal: {GOAL}
[Documentation of the perception and control primitives follows.]

===== HOW THIS SESSION WORKS =====
You are the coder in an automated, audited loop, working in this directory.
Each turn you write or revise the program as stage files (below); the loop
then executes it from scratch on a batch of scenes, an independent auditor
watches the videos, and you get a results digest: per scene, the task
outcome, the auditor's verdict on every stage and the auditor's account of
what happened. Nobody else reads your replies: the loop only looks for the
last word, READY or EXPERIMENT.

Scenes
- The run works through 15 fixed scenes with randomized object placements,
  in a pool that starts at 5 scenes. Every revision is run on a 2-scene
  batch that the loop picks (the scenes where the focus stage last failed
  come first). When the focus stage passes on both, the loop runs the other
  scenes of the pool with the same program. A stage counts as passed only
  when the auditor passes it on every scene of the pool. Once every stage
  passes with reward 1.0 on every scene of the pool, the same program is run
  on 5 new scenes and the pool grows 5 -> 10 -> 15; the run is accepted at
  15 of 15. Every scene run of your program counts against a fixed rollout
  budget. After the last rollout you still revise once
  more, and that revision is the final program.
- You debug one scene at a time, but you are writing ONE program for all
  scenes, and it must complete the ENTIRE task for any placement. Never
  write scene-specific code: no branch on a scene id, no constant read off
  one scene's placement, no strategy that only makes sense for the layout
  in front of you.

Plan
- Design your own high-level plan for the task as an ordered sequence of
  stages and write it to plan.md. Then for every stage write (a) what it
  does, concretely, for this task and robot, (b) the change in the OBJECTS'
  state that proves the stage succeeded, (c) how the verifier will observe
  that change. A stage is done when the object is where the plan says it
  should be, never because the robot reached some pose.
- Build the plan from the knowledge base first. For every object and fixture
  the task involves, and for every relation and state change the goal names,
  look up the entries whose object_class describes it and whose condition
  holds in this scene. The affordance_constraints decide what each stage may
  attempt, the spatial_relationships decide the approach, the
  effects_dependencies decide the order and what each stage's verifier must
  check. Cite the entries you used by their object_class.
- Before implementing, MEASURE. Run a measurement probe that segments each
  object with the perception APIs, deprojects it into a point cloud and
  measures every quantity the plan will depend on -- above all the
  open_quantities of the knowledge-base entries you rely on -- and record
  the results in measurements.json. When a later observation contradicts a
  recorded value, correct measurements.json and the plan promptly.

Stage files
- Write one file per stage: stages/01_<a>.py, stages/02_<b>.py, ... They are
  executed strictly in order in ONE shared namespace.
- Each stage file must end with exactly one call stage_done(ok, evidence),
  where ok is the verdict of the stage's own verifier: re-detect the object
  the task names from a fresh observation and check it is in the state the
  stage should leave it. Robot readings alone (gripper opening, EE pose) are
  never evidence that a stage succeeded. If ok is False the rollout stops
  there -- later stages do not run.

Audit and focus stage
- An independent auditor reviews every verdict against the rollout videos: a
  PASS the video does not support counts as FAIL, and a scene without task
  success is a failure whatever your verifiers reported.
- Each turn the feedback names the focus stage: the earliest stage that did
  not PASS on the scenes that ran. Work on that stage. If the evidence shows
  an earlier stage is the real cause you may revise that one instead.

Experiments
- Experiment before you fix whenever you are not sure of the cause.
- An experiment is a short program on one scene that does the minimum needed
  to reach the situation in question. Write experiments/NN_<slug>/hypothesis.md
  with these lines first:
      Hypothesis: <what you believe causes the failure>
      Prediction: <what the probe will show if it is true, and if it is false>
      Prefix: <a -> b>   (optional: stage names to run first)
      Scenes: <scene id(s) to run on>
  and experiments/NN_<slug>/experiment.py: plain code over the same APIs.
  Keep it small -- a handful of primitive calls, one question per experiment.
  End the reply with the single word EXPERIMENT. Record the outcome in
  diagnosis.md -- prediction held or refuted -- before acting on it.
  Experiments never change stage verdicts.

Lab notebook
- Keep diagnosis.md as a running lab notebook, appending one entry per turn.
  Never rewrite or condense earlier entries.
- After each results digest and before writing any code: check every
  hypothesis from your previous entry against the new evidence and mark it
  supported or refuted. Then record which stage failed on which scenes; the
  diagnosis, confirmed in BOTH the full log and video frames you extracted
  and looked at yourself; the hypotheses about the cause; and either the
  fix, with the concrete prediction it makes for the next rollout, or the
  experiment that will decide between the hypotheses.

===== READ FIRST: UNDERSTAND THE ENVIRONMENT AND THE SCENE =====
- setup_image.png: one training scene one second after the reset, the
  agentview camera on the left half and the wrist camera on the right half.
- libero_knowledge.json: the knowledge base, and the most important guiding
  file here. Its knowledge list holds entries of three types --
  affordance_constraints (what a shape affords, and what it forbids),
  spatial_relationships (how objects sit relative to each other, the
  fixtures and the robot) and effects_dependencies (what an action really
  changes, and what must hold before it works). Every sentence of an entry's
  knowledge is tagged [demonstrated], [inferred] or [untested] -- an
  [untested] sentence is a hypothesis for an experiment, not a rule.
  strategies names the approach; open_quantities lists what you must
  measure per scene, because the base records no numbers.

Your first turn, in this order:
1. Look at setup_image.png and find the knowledge-base entries that bear on
   THIS task: the objects involved, the strategies you will use, the effects
   and dependencies that order the stages, and every open_quantity those
   entries say must be measured.
2. Write the measurement probe experiments/00_measure/ and end your reply
   with the single word EXPERIMENT.
3. Fill measurements.json from the probe's output.
4. Open diagnosis.md with "Entry 0 -- my understanding of the environment and
   the scene", now with your measured numbers.
5. Then write plan.md and the stage files from those measurements, and reply
   READY.
\end{lstlisting}

\subsection{Auditor Prompt}
\label{sec:supp_prompt_auditor}

\Cref{lst:prompt_auditor} shows the auditor prompt. Placeholders carry the task goal, the stage plan, the stages already verified in earlier turns, and the execution digest of the current turn. The auditor returns per-stage verdicts and a time-ordered note for every scene.

\begin{lstlisting}[style=prompt,caption={Auditor prompt.},label={lst:prompt_auditor}]
You are an independent auditor for a robot-programming loop.

THE TASK
Goal (the robot's instruction, verbatim): {goal}
Success is decided by the simulator's own predicate on the TARGET object's
final pose; the env reward (1.0 / 0.0) in the results below is that signal.
A scene with reward 0 has at least one failing stage.

BEFORE JUDGING
Read libero_knowledge.json -- the knowledge base: how each kind of object
affords grasping and placing, how objects sit relative to fixtures and the
robot, and what an action really changes. Use it to recognise what you see
in the videos. Do not read the coder's own measurements -- they are not
needed to judge the videos.

WHAT THE CODER WROTE
A coder wrote a staged program -- {plan} -- one file per stage. After each
stage the program prints its own verdict, [STAGE <name>] PASS <evidence> or
FAIL <evidence>, and a FAIL stops the rollout.
{plan_block}{verified_block}
YOUR JOB
Decide, for EVERY scene and EVERY stage, whether that stage really
succeeded, and describe what actually happened so the coder can fix it. The
program's printed verdicts are the coder's CLAIMS, not evidence: a PASS may
be false, a FAIL may give a wrong reason. Judge from what actually happened:
the rollout videos and the measurements in the logs. Every video frame is
two views side by side: the agentview camera on the LEFT and the wrist
camera on the RIGHT. Base each verdict and each note on what the two views
show combined. A stage passes on a scene only if the TARGET object ends up
in the state that stage is supposed to produce. Robot readings alone
(gripper opening, end-effector pose) prove nothing. A stage that never ran
or was skipped fails.

For every scene write a note: a careful, factual account of what the video
shows, in time order, with the frame times you looked at.
Do not fix anything and do not propose fixes.

Turn {turn}. Scenes that ran and their results:
{digest}

Write your verdicts as a JSON list, one entry per scene, every stage:
  [{"seed": <int>, "note": "<what happened, in time order, with frame times>",
    "stages": {"<stage>": {"pass": true|false, "evidence": "<observed facts>"}, ...}}, ...]
\end{lstlisting}

\subsection{Per-Turn Feedback and Experiment Protocol}
\label{sec:supp_prompt_feedback}

Each turn, the coder receives the execution digest, the auditor grid with notes, the rollout budget, and a closing instruction. \Cref{lst:feedback_example} shows the feedback from an early turn of the LIBERO-PRO Goal task that places the plate on the stove, with the auditor note abridged. When the coder answers with an experiment, the loop runs it and returns its log and video together with the instruction in \cref{lst:prompt_exp_tail}. \Cref{lst:hypothesis_example} shows the hypothesis file the coder wrote for its first diagnostic experiment on the same task, which tests whether a reach limit or a collision prevented the gripper from reaching the plate rim.

\begin{lstlisting}[style=prompt,caption={Per-turn feedback from an early turn on the plate-on-stove task.},label={lst:feedback_example}]
Your program.py was executed FROM SCRATCH on 2 scenes (randomized object
placements). It succeeded on 0 of 2. Per-scene results:
  seed 110000: FAILURE (reward 0.000)
      [STAGE grasp] FAIL plate dropped during test lift: grip=0.0146
  seed 110001: FAILURE (reward 0.000)
      [STAGE grasp] FAIL plate dropped during test lift: grip=0.0132

An independent auditor checked your verdicts against the rollout videos.
The auditor's per-stage verdicts (these are ground truth; a PASS of yours
that the auditor marks fail counts as FAIL):
  scene 110000 (reward 0.0): grasp: FAIL  place: FAIL
    note: t=0.0s (f001): Initial scene. Agentview: plate (white with red
          concentric ring rim) flat on the table in the lower-center area;
          gray patterned bowl above and slightly right of the plate; [...]
          t=1.2s (f007): Agentview: arm extended far forward, hand/gripper
          low near the table surface beside the plate. Wrist view: the plate
          fills the right side of the frame; the finger pads are visible at
          the bottom of the wrist view, but the plate rim sits clearly ABOVE
          the pad level -- the pads have not reached the rim height. t=1.6s
          (f009): gripper is closed near the plate on the table; the fingers
          appear to have caught the very top edge of the rim but the hold is
          marginal. t=2.0s (f011): the plate appears tipped up onto its edge
          momentarily as the fingers tried to lift it, indicating it slipped
          out during the test lift. Log confirms: pad_bot=0.0174 vs
          rim_top=0.0069 (pads 10.5mm above the rim); grip after close was
          0.0474 (marginal contact with rim top edge), grip after test lift
          was 0.0146 (plate slipped out).
  scene 110001 (reward 0.0): grasp: FAIL  place: FAIL
    note: [...]

Rollout budget: 4 of 45 scene runs used, 41 left.

Focus stage: grasp
[closing instruction as above]
\end{lstlisting}

\begin{lstlisting}[style=prompt,caption={Instruction returned with each experiment result.},label={lst:prompt_exp_tail}]
Append to diagnosis.md: the hypothesis, the prediction, what was actually
observed (cite log lines and video frames you looked at), and whether the
prediction held or was refuted. Then continue with the focus stage: revise
it and end your reply with READY, or write another experiment and end with
EXPERIMENT. Experiments do not change any stage verdict.
\end{lstlisting}

\begin{lstlisting}[style=prompt,caption={Hypothesis file of the coder's first diagnostic experiment on the plate-on-stove task.},label={lst:hypothesis_example}]
Hypothesis: The arm can't reach the plate rim at x~0.71 (pad bottom z=0.015,
  rim top z=0.007 -> 8mm gap). Straddling from the -x direction (x~0.65,
  6.5cm closer to robot) should allow the arm to reach 3-5cm lower, placing
  the pads at the rim level. A diagonal (135 deg) straddle avoids the bowl
  collision while bringing the gripper 4.6cm closer.
Prediction: At the 135 deg straddle point (x~0.668), the pad bottom z will be
  at or below 0.007 (the rim top). A grip after close will hold during a 5cm
  test lift. At the 90 deg straddle (x=0.714, current approach), the pad
  bottom remains above 0.010. At the 180 deg straddle (x=0.649), reach is
  best but the bowl may be displaced.
Prefix:
Scenes: 110000
\end{lstlisting}
\twocolumn

\FloatBarrier
\section{Real-Robot Experimental Details}
\label{sec:supp_real_robot}

\begin{figure*}[t]
  \centering
  \includegraphics[width=\linewidth]{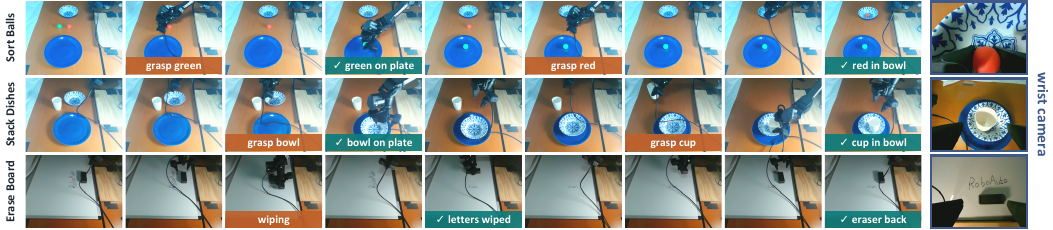}
  \caption{\textbf{Real-robot rollouts in detail.} RoboAuto programs on a WidowX AI arm for three tasks. Each row shows nine scene-camera keyframes in time order and one wrist-camera frame. Orange labels mark actions in progress, and checked labels mark stages verified before the next stage begins.}
  \label{fig:supp_real_robot_long}
\end{figure*}

\paragraph{Hardware setup}
We conduct real-robot experiments using a Trossen Robotics WidowX AI arm equipped with an Intel RealSense D405 RGB-D camera and a Logitech C920 RGB camera. The cameras provide complementary geometric and visual observations of the robot, objects, and workspace. All methods use the same hardware, sensing configuration, and control interface.

\paragraph{Knowledge base}
For the real-robot experiments, RoboAuto directly reuses its hierarchical knowledge base acquired from LIBERO tasks, which combines reusable object-centric physical knowledge with executable skills. All real-robot tasks are held out and distinct from those used for knowledge-base construction. For a fair comparison, ASPIRE also transfers its skill library acquired from the LIBERO tasks. Despite the domain shift from simulation to the physical world, RoboAuto effectively retrieves, grounds, and composes its accumulated knowledge for previously unseen tasks and configurations. Its substantial advantage over ASPIRE shows that the hierarchical knowledge base acquired in simulation transfers to unseen real-robot tasks.

\paragraph{Experimental setup}
We use Opus~5 with a 1M-token context window for the search component of both program-search methods, RoboAuto and ASPIRE, and allocate each a budget of 10 search rollouts to ensure a controlled comparison. Because no simulator success predicate is available on the real robot, a human operator judges whether each search rollout succeeds; this judgment replaces the environment reward used in simulation and is returned to the search loop. The operator also resets the scene after each rollout. For both search and evaluation, initial object positions are randomly sampled within a fixed reset region, and the positions used for evaluation are sampled independently of and differ from those used during search. We evaluate RoboAuto, $\pi_{0.5}$, and ASPIRE on three long-horizon manipulation tasks using a WidowX AI arm. In \textbf{Sort Balls}, the robot must place a green ping-pong ball on a plate and a red ping-pong ball in a bowl. In \textbf{Stack Dishes}, it must place a low-rimmed bowl on a plate and then place a cup upright in the bowl. In \textbf{Erase Board}, it must pick up a compliant eraser, erase ``RoboAuto'' from a whiteboard, and return the eraser to its original position. Each method is evaluated over 20 trials per task with initial object positions drawn from the same reset region but not used during search. After each trial, a human evaluator assigns a single stage-completion score, the fraction of stages completed, and performance is the mean score over trials.

\paragraph{Task evaluation}
We evaluate performance based on stage completion rather than binary success on the full long-horizon task. \textbf{Sort Balls} consists of four stages: (1) picking up the green ping-pong ball, (2) placing it on the plate, (3) picking up the red ping-pong ball, and (4) placing it in the bowl. \textbf{Stack Dishes} also consists of four stages: (1) picking up the bowl, (2) placing the bowl on the plate, (3) picking up the cup, and (4) placing the cup upright in the bowl. Each completed stage in these two tasks contributes one quarter of the total score. \textbf{Erase Board} consists of three stages: (1) picking up the eraser, (2) erasing the eight letters in ``RoboAuto,'' and (3) returning the eraser to its original position. Each stage contributes one third of the total score. Because the erasing stage involves eight letters, we assign proportional credit based on the number of letters successfully erased: erasing $n$ letters yields $n/8$ of the score allocated to this stage. This stage-based metric captures meaningful partial progress and provides a more informative evaluation than measuring only whether the entire long-horizon task is completed.


\end{document}